\documentclass[10pt,twocolumn,letterpaper]{article}

\usepackage[final]{cvpr}      

\usepackage{graphicx}
\usepackage{amsmath}
\usepackage{amssymb}
\usepackage{booktabs}
\usepackage{xcolor}
\usepackage{appendix}
\usepackage{multirow}

\usepackage[ruled, noline]{algorithm2e}

\usepackage[pagebackref,breaklinks,colorlinks]{hyperref}

\usepackage[capitalize]{cleveref}
\crefname{section}{Sec.}{Secs.}
\Crefname{section}{Section}{Sections}
\Crefname{table}{Table}{Tables}
\crefname{table}{Tab.}{Tabs.}

\def\cvprPaperID{****} 
\def\confName{CVPR}
\def\confYear{2023}

\begin{document}

\title{Traditional Classification Neural Networks are Good Generators: They are Competitive with DDPMs and GANs}

\author{Guangrun Wang$^{1}$ \quad Philip H.S. Torr$^1$ \\
\small$^1$ University of Oxford \quad
\\{\tt\small \{guangrun.wang,philip.torr\}@eng.ox.ac.uk}
}


\twocolumn[{%
\renewcommand\twocolumn[1][]{#1}%
\vspace{-1em}
\maketitle
\vspace{-1em}
\begin{center}
    \centering
    \vspace{-0.1in}
    \includegraphics[width=1.0\linewidth]{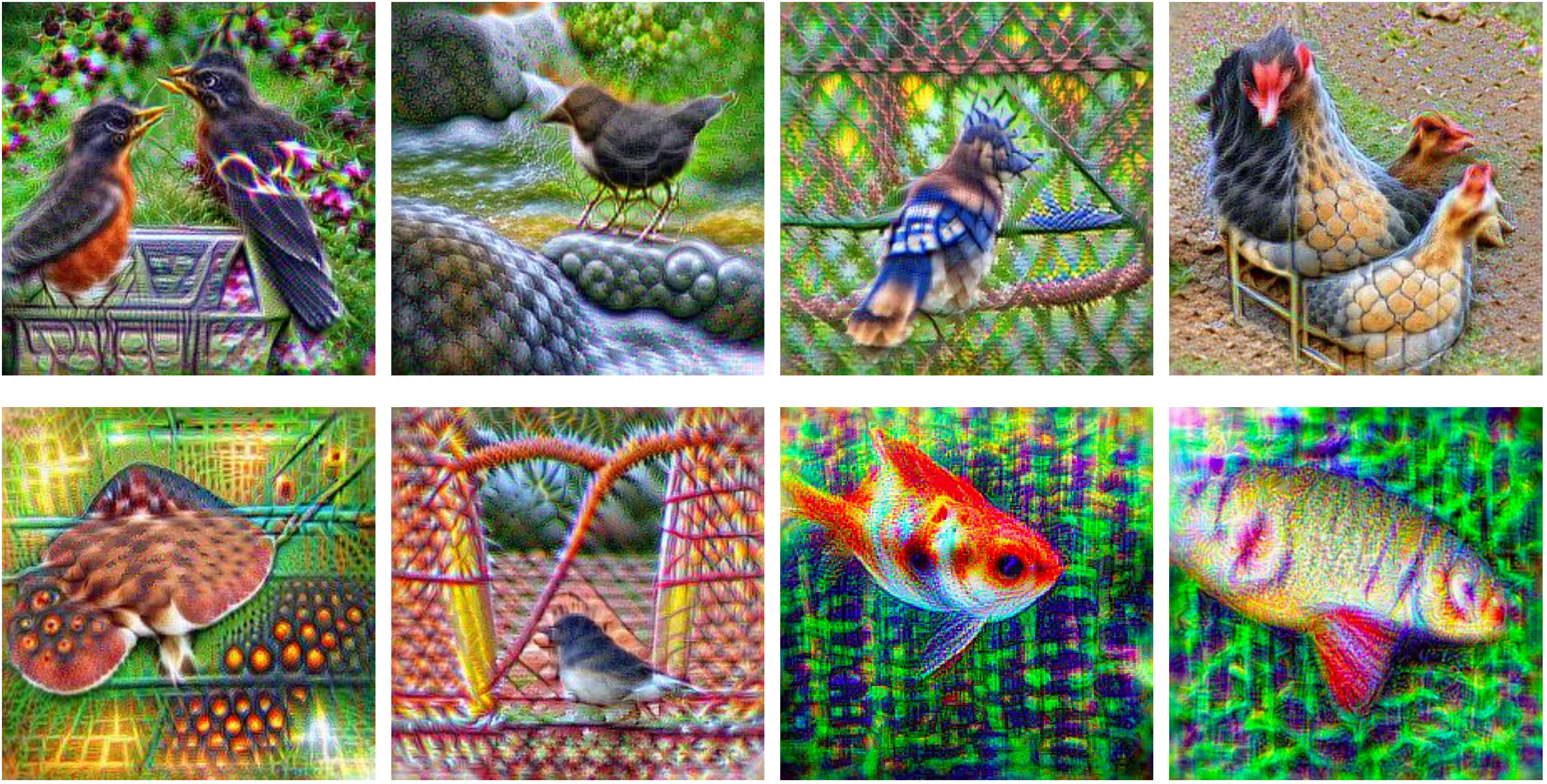}
    \captionof{figure}{Selected samples generated by classification neural networks on the ImageNet 256$\times$256 task (More samples are in the appendix.).}
    \label{fig:intro}
\end{center}%
}]

\begin{abstract}
Classifiers and generators have long been separated. We break down this separation and showcase that conventional neural network classifiers can generate high-quality images of a large number of categories, being comparable to the state-of-the-art generative models (e.g., DDPMs and GANs).
We achieve this by computing the partial derivative of the classification loss function with respect to the input to optimize the input to produce an image. Since it is widely known that directly optimizing the inputs is similar to targeted adversarial attacks incapable of generating human-meaningful images, we propose a mask-based stochastic reconstruction module to make the gradients semantic-aware to synthesize plausible images. We further propose a progressive-resolution technique to guarantee fidelity, which produces photorealistic images. Furthermore, we introduce a distance metric loss and a non-trivial distribution loss to ensure classification neural networks can synthesize diverse and high-fidelity images. Using traditional neural network classifiers, we can generate good-quality images of 256$\times$256 resolution on ImageNet. Intriguingly, our method is also applicable to text-to-image generation by regarding image-text foundation models as generalized classifiers.

Proving that classifiers have learned the data distribution and are ready for image generation has far-reaching implications, for classifiers are much easier to train than generative models like DDPMs and GANs. We don't even need to train classification models because tons of public ones are available for download. Also, this holds great potential for the interpretability and robustness of classifiers. Project page is at \url{https://classifier-as-generator.github.io/}.
\end{abstract}


\section{Introduction}

Neural network classifiers \cite{lecun1998gradient,krizhevsky2017imagenet,simonyan2014very,he2016deep,dosovitskiy2020image} have gained remarkable progress in computer vision and machine learning. They achieve human-level performance in many tasks.

Lagging behind neural network classifiers are neural network generators. Over the past few years, neural network generators have made some progress, e.g., in generating texts \cite{brown2020language}, images \cite{ramesh2021zero,ramesh2022hierarchical,rombach2022high,saharia2022photorealistic}, sounds \cite{oord2016wavenet,dhariwal2020jukebox}, videos \cite{singer2022make,villegas2022phenaki,zhuo2022fast}, lights \cite{chen2022text2light,wang2022stylelight}, and 3D scenes \cite{chan2022efficient,chan2021pi}. 
Despite that, they are still inferior to neural network classifiers. 
On the one hand, neural network classifiers are easier to learn than generators in terms of efficiency and stability. For example, training a classification net is straightforward; but training a GAN \cite{goodfellow2020generative} could easily lead to model collapse and non-convergence, and training a DDPM \cite{ho2020denoising,dhariwal2021diffusion,nichol2021improved} requires complicated efforts.
On the other hand, neural network classifiers can better model the data's true distribution than generators. 
Specifically, classifiers can project the data into the correct latent space so that objects can be reasonably distributed in the latent space according to the category they belong to; in contrast, the mapping between the latent space and the data space for GANs and DDPMs is chaotic, i.e., their latent codes contain less semantic signals, especially for DDPMs.
As a typical consequence, the features learned by neural network generators are often less generalizable and transferrable than classifiers.

\begin{figure}
\centering
\includegraphics[width=1.\linewidth]{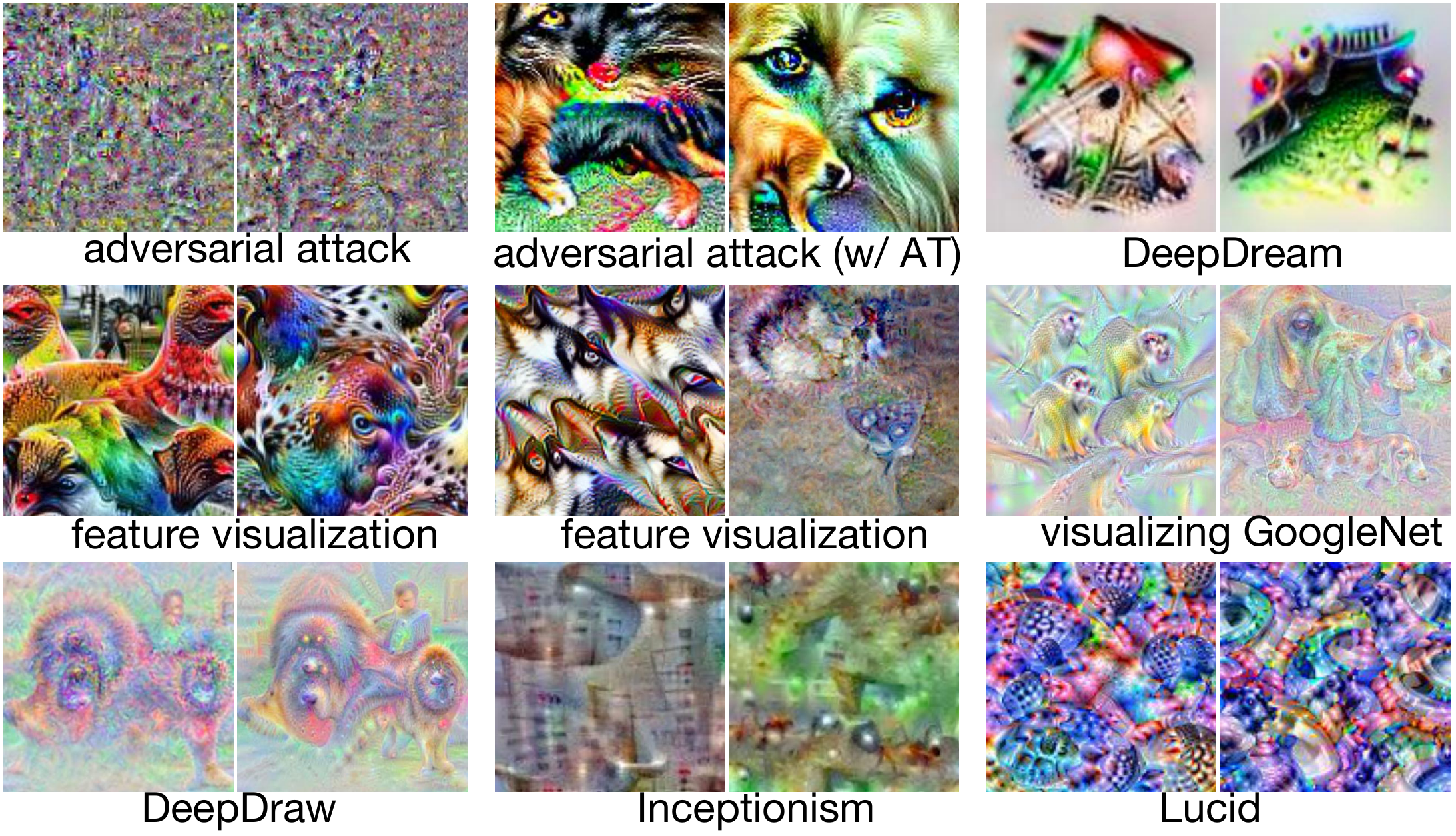}
\vspace{-11pt}
\caption{Samples of feature visualization methods.
These methods include adversarial attack \cite{goodfellow2014explaining}, adversarial robustness as a prior \cite{engstrom2019adversarial}, DeepDream \cite{Mordvintsev2016DeepDream}, feature visualization \cite{Olah2017Feature}, reproduced feature visualization \cite{engstrom2019adversarial}, visualizing GoogleNet \cite{oygard2015Visualizing}, DeepDraw \cite{oygard2015DeepDraw}, Inceptionism \cite{mordvintsev2015inceptionism}, and Lucid \cite{Olah2017Lucid,engstrom2019adversarial}.
Here, AT denotes adversarial training.
Note that these samples are from their papers or blogs. Thank the authors.}\label{fig:feature_vis}
\vspace{-11pt}
\end{figure}

Since neural network classifiers are so knowledgeable and data-distribution-aware, we ask: are they ready for image generation?
It is widely known that directly optimizing the inputs of neural network classifiers does not produce human-meaningful images, for it is equivalent to a targeted adversarial attack.
Given this, prior arts utilize multiple regularization methods to help visualize the features of neural networks, e.g., via frequency penalization \cite{mahendran2015understanding,nguyen2015deep,oygard2015Visualizing,Tyka2016Class,Mordvintsev2016DeepDream,oygard2015DeepDraw}, transformation invariance \cite{mordvintsev2015inceptionism,oygard2015Visualizing,Tyka2016Class,Mordvintsev2016DeepDream,Olah2017Feature,oygard2015DeepDraw,Olah2017Lucid}, adversarial training \cite{santurkar2019image,engstrom2019adversarial,tsipras2018robustness}, and energy-model-based training \cite{grathwohl2019your}.
However, as tools for feature visualization, while these methods can certainly produce some better visualizations than random noise, there is a considerable gap between their visualizations and realistic images in terms of plausibility, fidelity, and diversity (see Figure \ref{fig:feature_vis}).
\textbf{First}, their visualizations unreasonably contain duplicated patterns everywhere (e.g., many copies of dog eyes in an image), which looks quite implausible.
\textbf{Second}, their visualizations lack fidelity. The produced images are low-quality, looking fake and unnatural.
\textbf{Third}, their visualizations are severely lacking diversity.
Visualizations produced by different random noises share similar content and style. Besides, some method needs well adversarial training that often suffers from robust overfitting and is time-consuming.

In this paper, we showcase that traditional classification neural networks can synthesize high-quality images at ImageNet large scale that contains a thousand object categories, being comparable to state-of-the-art generative models (e.g., DDPMs and GANs).
To accomplish this, we compute the partial derivatives of the classification loss function with respect to the inputs to optimize the inputs to produce images.
But this alone cannot generate plausible, realistic, and diverse images.
For \textbf{plausibility}, we introduce a mask-based stochastic reconstruction module to make the gradients semantic-aware to generate reasonable images. This module can significantly improve plausibility as well as fidelity.
For \textbf{fidelity}, we further propose a progressive-resolution generation technique, starting with optimizing 64$\times$64 images and ending up with optimizing 256$\times$256 images.
This technique could reduce diversity as it is widely known that fidelity and diversity have a tradeoff.
Finally, to break the curse of compromise between fidelity and \textbf{diversity}, we introduce a distance metric loss as well as a nontrivial distribution loss to guarantee the synthesized images are of good diversity.

Intriguingly, our method also applies to text-to-image generation, for we can regard image-text foundation models as generalized classifiers -- maximizing the inner product between an image and a text reduces to classifying this image to this text. Our method can use texts to produce meaningful samples.

In summary, this paper makes five contributions.\begin{itemize}
\item We showcase that traditional neural networks can generate high-quality images at ImageNet large scale, competitive with state-of-the-are generative models.
This has far-reaching implications since classifiers are easier to obtain than generators; it also holds potential for the interpretability and robustness of classifiers.
\item We address the implausibility of directly optimizing the inputs using a mask-based stochastic reconstruction module to make the gradients semantic-aware to synthesize reasonable images.
\item We propose a progressive-resolution generation technique to guarantee fidelity, which is capable of producing photorealistic images.
\item To increase diversity, we introduce a distance metric loss and a nontrivial distribution loss to guarantee the diversity of the generated images of the same class.
\item Using traditional neural network classifiers, we can generate good-quality samples on the ImageNet 256$\times$256 task. Intriguingly, our method can also achieve text-to-image generation via regarding image-text foundation models as generalized classifiers.
\end{itemize}

\begin{figure*}
\centering
\includegraphics[width=0.85\linewidth]{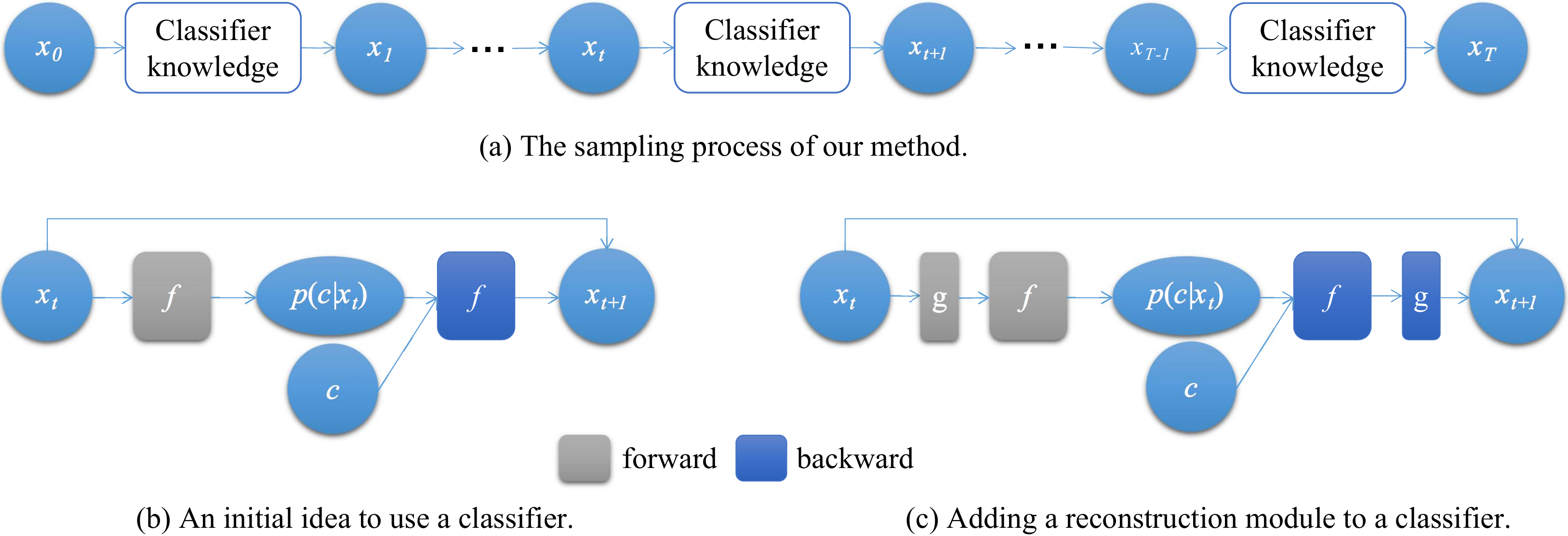}
\vspace{-5pt}
\caption{Illustration of our method. (a) is the sampling process of our method, and the key to its success is to exploit the knowledge of the classifier, which can be implemented in two ways, as illustrated in Figures (b) and (c). 
(b) An initial idea to use a classifier for generation, but the gradients are, unfortunately, semantic-agnostic. (c) By adding a mask-based stochastic reconstruction module to the classifier, the gradients are made semantic-aware to synthesize human-meaningful images.}\label{fig:method}
\vspace{-11pt}
\end{figure*}

\section{Related Work}

\textbf{Feature visualization.} Visualizing neural network features has been a long-standing vision \cite{yosinski2015understanding}, where the focus is on understanding neural networks rather than generating images.
To produce human-meaningful visualizations, prior works use a variety of regularization methods to make the results somewhat interpretable.
These regularization methods include high-frequency penalty \cite{mahendran2015understanding,nguyen2015deep,oygard2015Visualizing,Tyka2016Class,Mordvintsev2016DeepDream,oygard2015DeepDraw}, transformation invariance \cite{mordvintsev2015inceptionism,oygard2015Visualizing,Tyka2016Class,Mordvintsev2016DeepDream,Olah2017Feature,oygard2015DeepDraw, Olah2017Lucid}, adversarial training \cite{santurkar2019image,engstrom2019adversarial,tsipras2018robustness}, and energy-model-based training \cite{grathwohl2019your}.
However, these methods are just limited to visualization for explainability, and the visualizations look fake regarding plausibility, fidelity, and diversity.
Classifiers acting as good generators are still absent currently.


\textbf{Classifier-guided generative models.} Some work has begun to use classifiers as guidance to control the production of generative models.
This involves using a pretrained classifier to optimize the latent space of GANs (often called GAN inversion) \cite{nguyen2016synthesizing,nguyen2017plug,galatolo2021generating,patashnik2021styleclip} and to optimize the sampling direction of DDPMs \cite{dhariwal2021diffusion,sohl2015deep,song2020score}.
While these methods demonstrate the power of classifiers for image generation, they are heavily dependent on existing generators (such as GANs or DDPMs).
Unlike these methods, we showcase that traditional neural network classifiers already know the data distribution and are ready for image generation.

\textbf{DDPMs.} Although significantly different from DDPMs, our method has some connections to DDPMs \cite{ho2020denoising,dhariwal2021diffusion,nichol2021improved}.
DDPMs are likelihood-based generative models that have been shown to produce high-fidelity images .
However, just as Rome wasn't built in a day, DDPMs are being improved through the joint efforts of many articles \cite{song2019generative,ho2020denoising,sohl2015deep,jolicoeur2020adversarial,song2020score,song2020denoising,nichol2021improved,saharia2022image,alias2017variational}.
While DDPMs have achieved some success, they have some deficiencies over neural network classifiers.
First, DDPMs are harder to train than neural network classifiers.
Second, the latent space of DDPMs is less semantic than that of neural network classifiers.

\section{Neural Network Classifier as Generator}\label{sec:method}

In this section, we first review neural network classifiers.
Then, we introduce how to convert their learned knowledge into reasonable, high-fidelity, and diverse images from three aspects: plausibility, fidelity, and diversity. Finally, we present the connection between our method and DDPMs, and how to extend it for text-to-image generation.
For the sake of brevity, we use CaG to term our method, which is short for ``Classifier as Generator."

\subsection{Neural Network Classifier}\label{sec:cls}

The formulation of neural network classifiers is very simple, and people are quite familiar with it. Let $f$ denote a neural network, $x$ its input image, and $p$ its output logits before softmax normalization. Then a neural network classifier is written as $p = f(x)$. The prediction of this classifier is the index in which the value is the largest in $p$. More generally, a neural network classifier can be a cross-modal model for a text-to-image modeling task like CLIP \cite{radford2021learning}.

Unlike all generators, neural network classifiers are directly trained independently on classification tasks such as ImageNet or CIFAR-10. 
Note that training neural network classifiers is much easier than generators. Also, tons of classification models are available for download online, so we don't even need to train them.

We hope that using the knowledge of the neural network classifier, through step-by-step iterations, we can turn a random input into an expected sample and achieve image generation.
Figure \ref{fig:method} (a) shows a framework we envision.

\subsubsection{An initial idea for generation}

With a trained neural network classifier $f$ and a given class label $c$, one can use a classification loss to encourage the classifier to predict this label. In this way, a generation problem reduces to an optimization problem.
Specifically, one can compute the partial derivative of the loss with respect to the input $x_t$ and optimize $x_t$ such that the prediction of the neural network classifier is $c$:\begin{equation}\label{eqn:gen}
x_{t+1} = x_t - \arg\mathop{\min}\limits_{\Delta x_t}\mathcal{L}_\mathrm{cls}(f(x_t + \Delta x_t), c),
\end{equation}where
$t$ denotes the time sequence of the optimization. $\mathcal{L}_\mathrm{cls}$ is the classification loss; $x_0$ is an initial tensor.

Unfortunately, Eqn. \eqref{eqn:gen} is equivalent to a targeted adversarial attack, which is widely known to be unable to generate human-meaningful images.
An initial idea for using a neural classifier as a generator is shown in Figure \ref{fig:method} (b).

\subsection{Mask-Based Stochastic Reconstruction Module}

Mathematically, the optimization problem in Eqn. \eqref{eqn:gen} treats the high-dimensional input space as the unknown variable to be solved, which has infinitely many solutions. Many of these solutions are semantic-agnostic.

To address the above dilemma, we propose a mask-based stochastic reconstruction model to make gradients semantic-aware to generate plausible images. Specifically, the gradients may guide the variables to be solved towards a local optimum to satisfy Eqn. \eqref{eqn:gen}, resulting in human-meaningless images. Fortunately, \cite{he2022masked} and \cite{wang2022semantic} showcase that mask-based stochastic reconstruction models are semantically aware. Inspired by these two papers, we add a mask-based random reconstruction module to the input images: $x \leftarrow g(x)$, where $g$ is a masked auto-encoder \cite{he2022masked}, defined by:\begin{equation}
g^* = \mathop{\min}\limits_{g}\mathbb{E}_{x\sim \mathcal{D}} \| m(g(\overline{m}(x))) - m(x) \|_2^2,
\end{equation}where $\mathbb{E}$ represents the mathematical expectation. $\mathcal{D}$ represents the data distribution.
$m$ represents an operation that obtains the masked region of an image.
$\overline{m}$ is complementary to $m$, meaning a procedure that gets the unmasked part of an image.
Hence, Eqn. \eqref{eqn:gen} can be rewritten as:\begin{equation}\label{eqn:mask}
x_{t+1} = x_t - \arg\mathop{\min}\limits_{\Delta x_t}\mathcal{L}_\mathrm{cls}(f(g(x_t + \Delta x_t)), c).
\end{equation}Here, $g$ is a general mask-based stochastic reconstruction module that can be pretrained with general images. Many of them are available for download online, so we don't need to train them, either (see the experimental section for detail).

An illustration of a classifier with a mask-based stochastic reconstruction module is presented in Figure \ref{fig:method} (c).
We explain why $g$ can make gradients semantic-aware as follows: in the absence of $g$, there are many gradient directions that can make the loss in Eqn \eqref{eqn:gen} drop; but in the presence of $g$, many of these gradient directions are invalid because they cannot complete the reconstruction task to reach the final classification goal;
in this way, the gradients are regularized to be semantic-aware of the objects to synthesize plausible images.

\subsection{Progressive-Resolution Generation Technique}

Empirically, we find that if the input resolution is larger, the generated images are more diverse but with some minor noise on the image surface. Although these slight noises do not affect humans' recognition of the images, they affect the fidelity of the images. On the contrary, if the input resolution is smaller, the resulting images will be less diverse, but the images will be more realistic.

To guarantee fidelity, we propose a progressive resolution generation technique.
We start by producing images of 64$\times$64 resolution and gradually increase the resolution of the resulting images exponentially.
Specifically, we sequentially generate images of 64$\times$64, 128$\times$128, and 256$\times$256.

This is definitely different from the traditional super-resolution task \footnote{In fact, our method lacks a super-resolution module commonly used in DDPMs or GANs, which puts our model at an unfair disadvantage.}; actually, it is a process similar to cognitive imagination. Initially, the overall framework configuration is imagined. The details are then gradually imagined.
Through this progressive process, we significantly improve the fidelity of the image.

\subsection{Distance Metric Loss}\label{sec:extra_loss}

On the scalability of our method, we ask: since we can impose a classification loss function to optimize the input to generate reasonable images, can we impose other loss functions to guide reasonable image generation?

In addition to fidelity, diversity is another critical metric for evaluating image generation quality.
Therefore, we introduce a diversity-encouraging loss function to verify the scalability of our method, hoping that it will improve the diversity of the generated images.
The diversity-encouraging loss function is defined as a distance metric loss:\begin{equation}\label{eqn:div}
\mathcal{L}_\mathrm{div} (x^1, \cdots, x^N)= \sum\limits_{\substack{i\neq j; \\ x^i, x^j \in y}} h^T(g(x^i))h(g(x^j)),
\end{equation}where $h$ is the feature extractor in the classifier $f$ in Eqn. \eqref{eqn:gen}, i.e., $h$ is the remaining part of $f$ after removing the last linear layer. $x^i$ indicates the $i$-th generated sample of class $y$. Here, $N$ represents the number of images generated for the same class $y$.
The meaning of Eqn. \eqref{eqn:div} is obvious, i.e., we hope the similarity between the images generated for the same class to be small. Combining Eqn. \eqref{eqn:mask} and \eqref{eqn:div} yields:\begin{equation}\label{eqn:final}
\begin{aligned}
x_{t+1}^i &= x_t^i - \arg\mathop{\min}\limits_{\Delta x_t^i} \Big(\mathcal{L}_\mathrm{cls}(f(g(x_t^i + \Delta x_t^i)), c)+ \\
& \mathcal{L}_\mathrm{div}(x_t^1,\cdots, x_t^i+\Delta x_t^i, \cdots, x_t^N)\Big).
\end{aligned}
\end{equation}

\textbf{Remark}. Experimental results show that the distance metric loss can improve the diversity of the generated images, also proving that our method has good scalability because maybe other loss functions for other tasks also work.

\begin{figure*}
\centering
\includegraphics[width=1.0\linewidth]{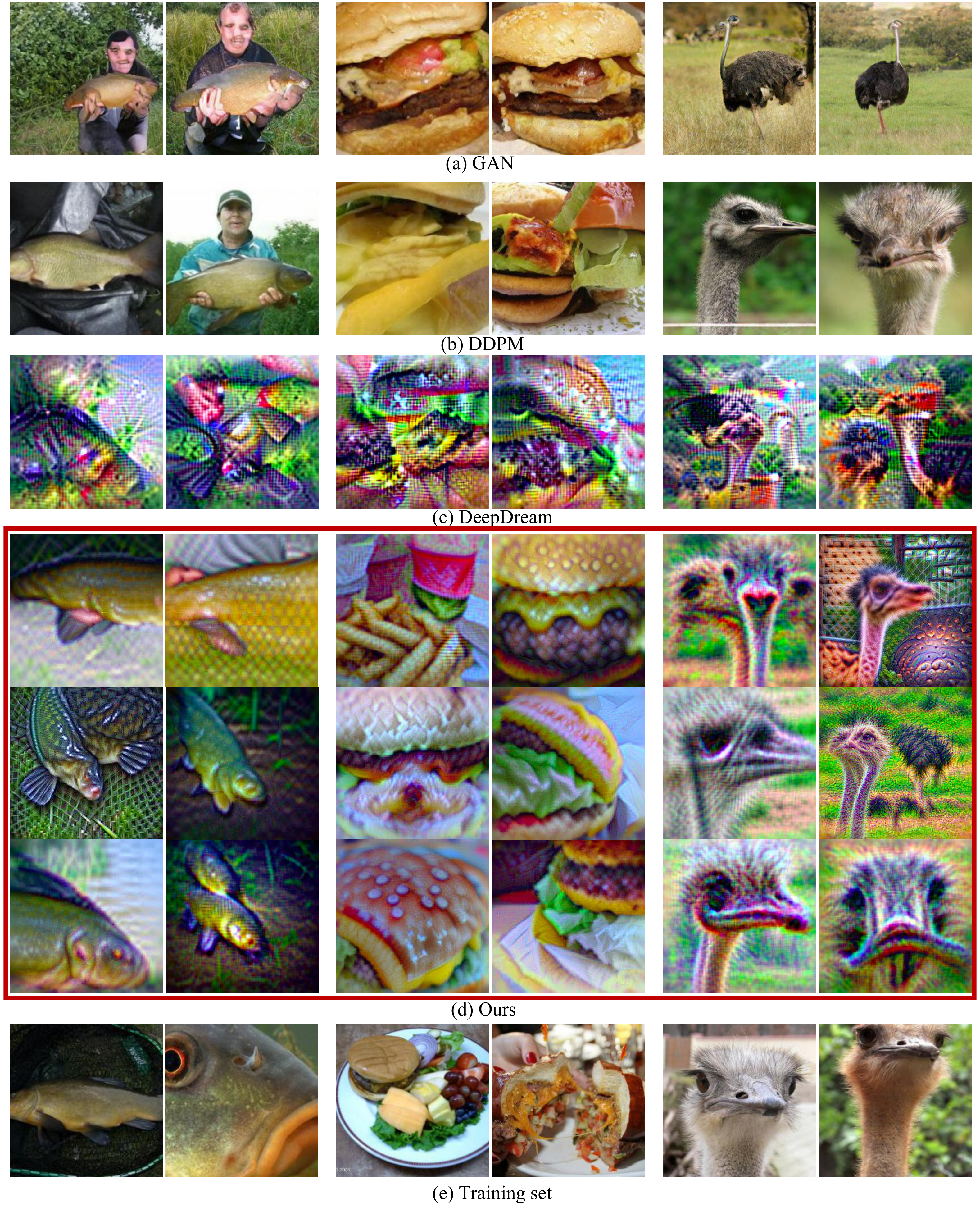}
\caption{Samples from a GAN (i.e., BigGAN-deep \cite{brock2018large}) vs. samples from a DDPM (i.e., ADM \cite{dhariwal2021diffusion}) vs. samples from a feature visualization method (i.e., DeepDream \cite{Mordvintsev2016DeepDream}) vs. samples from our method and samples from the training set. Note that some of these samples are from their paper. Thanks to the authors. More samples are in the appendix.}\label{fig:sota}
\end{figure*}

\begin{figure*}
\centering
\includegraphics[width=1.0\linewidth]{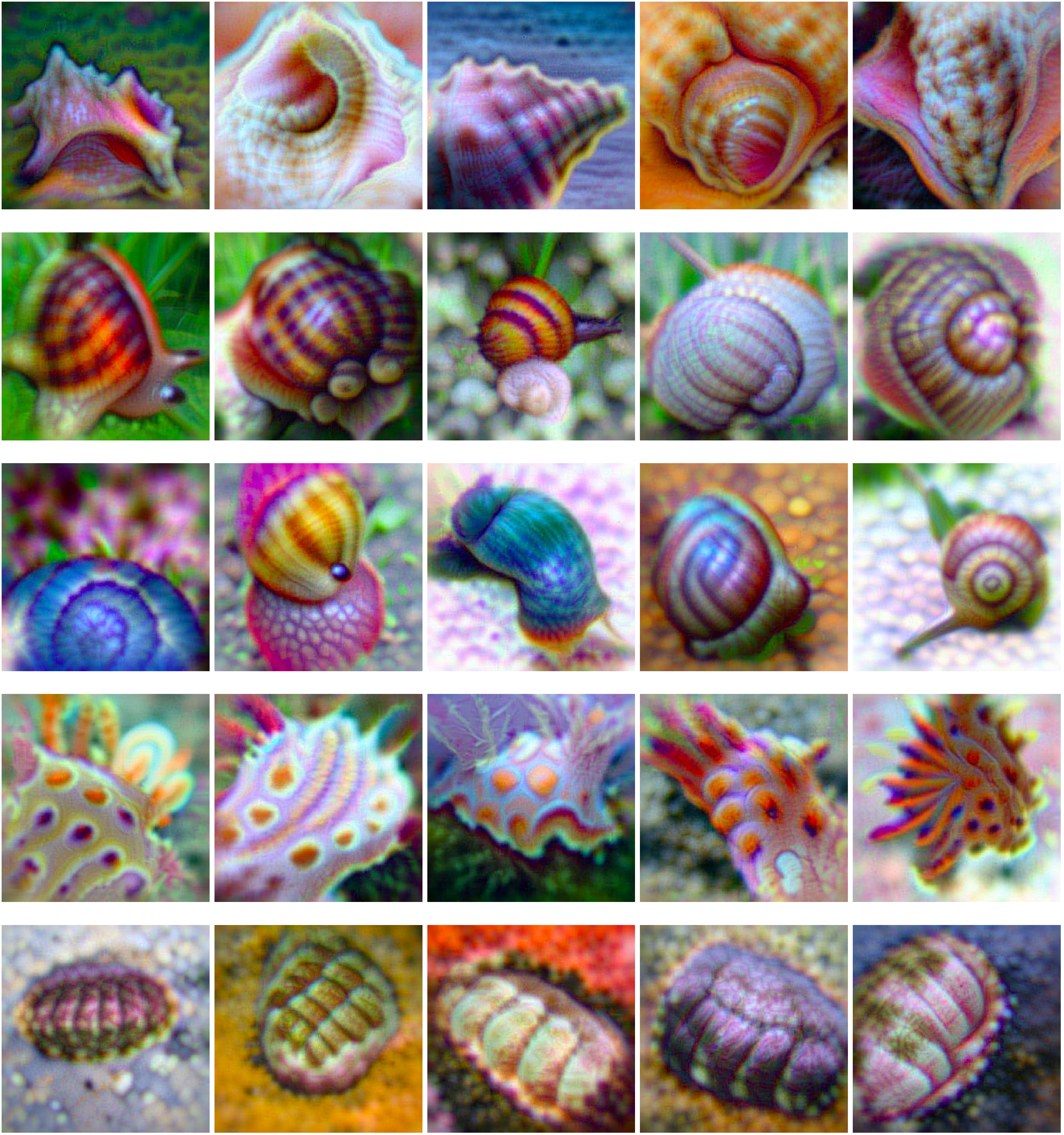}
\caption{Samples of CaG on ImageNet 256$\times$256. The classes are Class 112 (i.e., conch), Class 113 (i.e., snail), Class 113 (i.e., snail), Class 115 (i.e., sea slug), and Class 116 (i.e., chiton).
More samples are in the appendix.
}\label{fig:supp5}
\end{figure*}

\begin{figure*}
\centering
\includegraphics[width=1.0\linewidth]{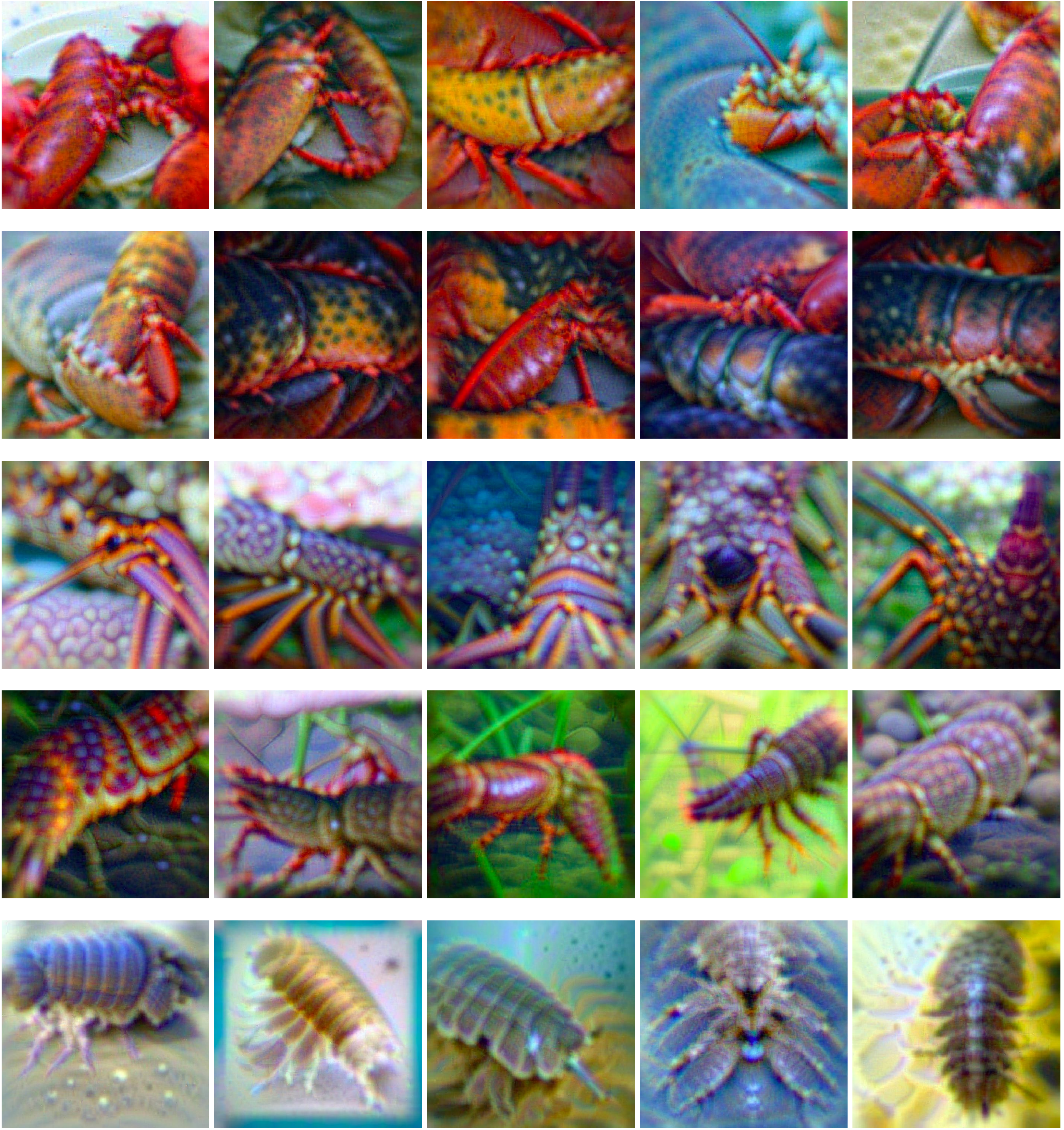}
\caption{Samples of CaG on ImageNet 256$\times$256. The classes are Class 117 (i.e., chambered nautilus), Class 118 (i.e., Dungeness crab), Class 120 (i.e., fiddler crab), Class 121 (i.e., king crab), and Class 125 (i.e., hermit crab).
More samples are in the appendix.
}\label{fig:supp7}
\end{figure*}

\begin{figure*}
\centering
\includegraphics[width=1.0\linewidth]{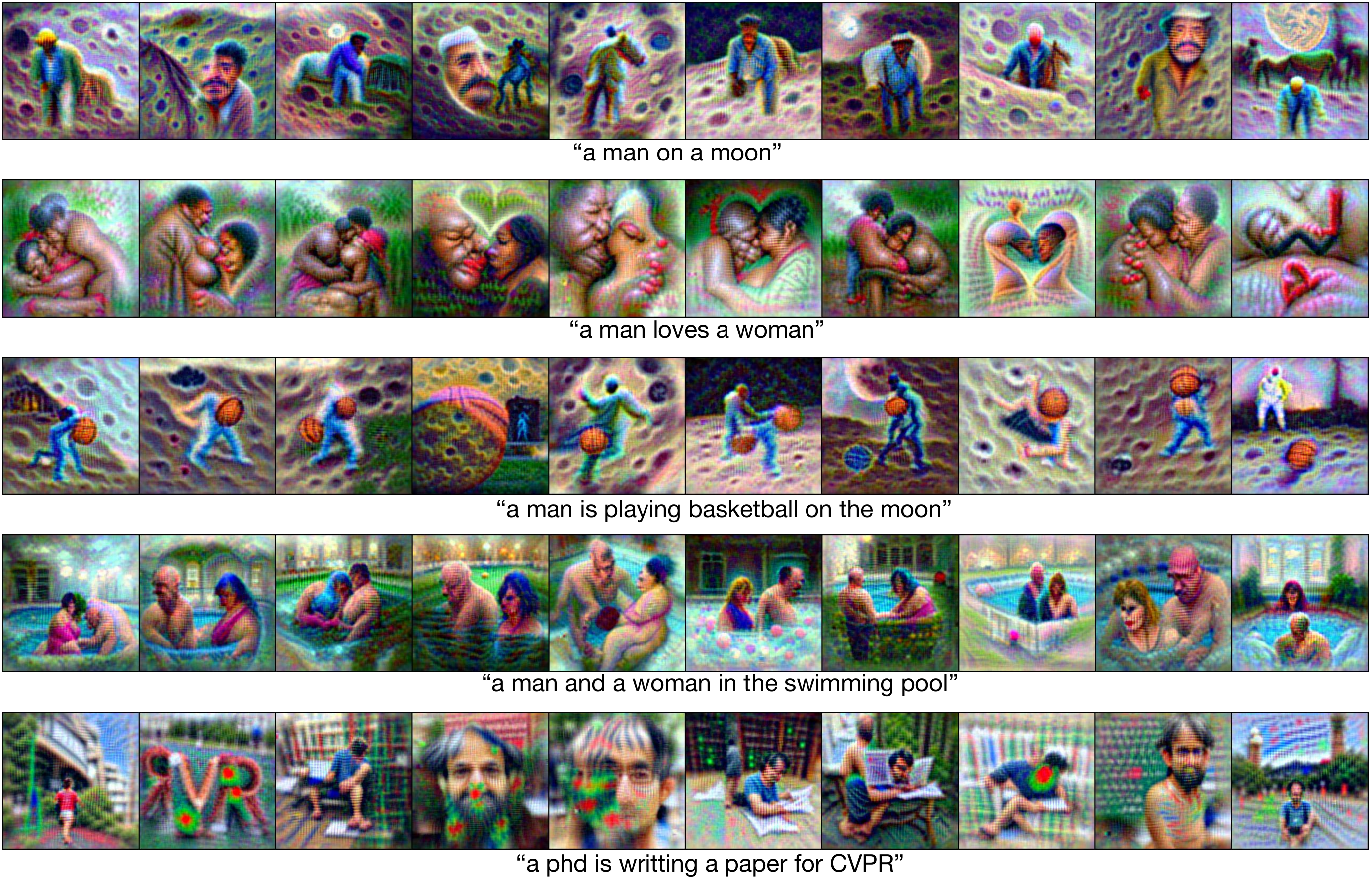}
\caption{Text-to-image generation using a pretrained text-image model CLIP \cite{radford2021learning}. Note that we are not chasing high sample quality here, but using fast optimization to get results. We also did not use distance metric loss to increase the diversity of samples. More samples are in the appendix.}\label{fig:t2i}
\end{figure*}

\begin{figure*}
\centering
\includegraphics[width=0.75\linewidth]{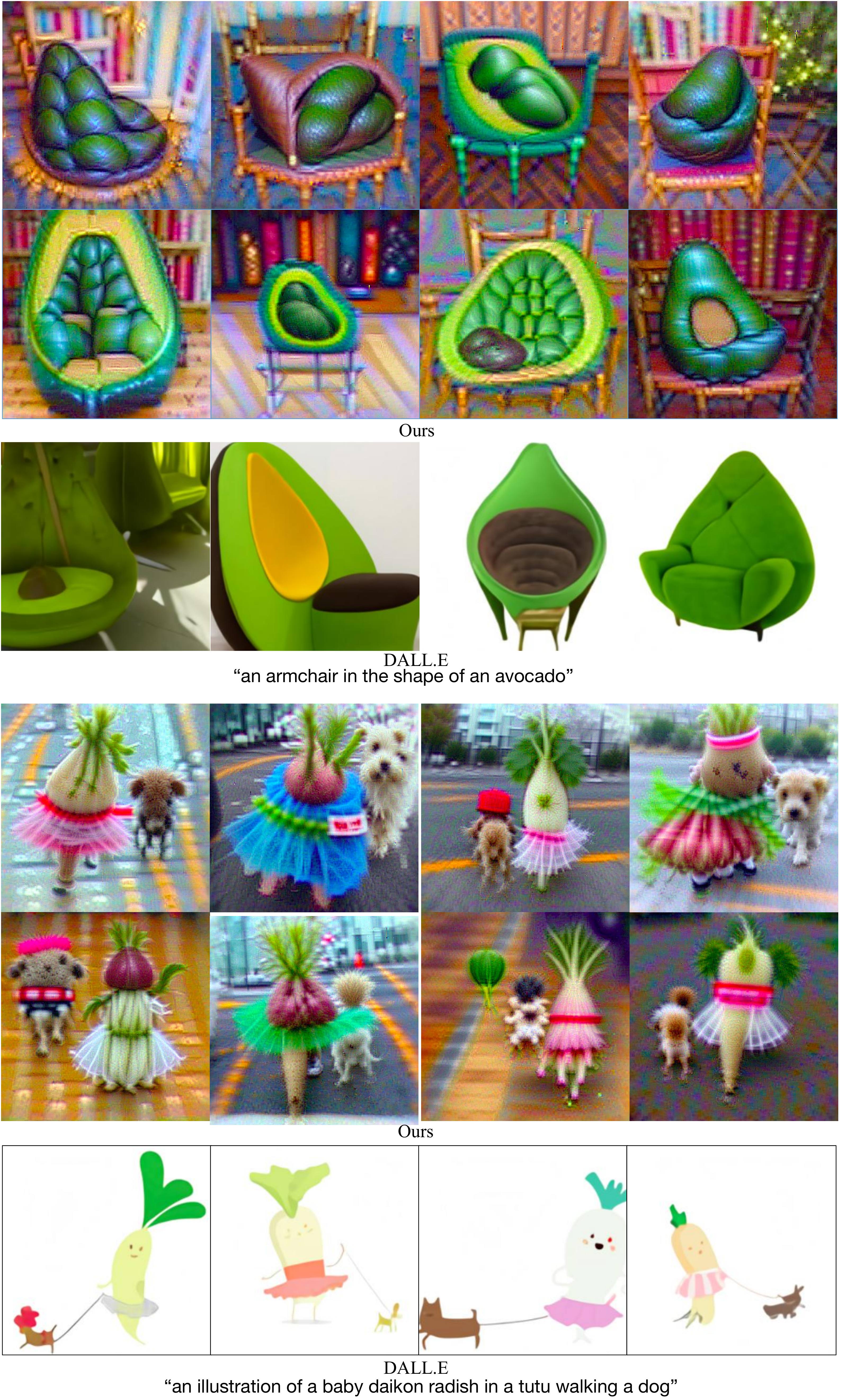}
\vspace{-6pt}
\caption{Text-to-image generation using a pretrained text-image model CLIP \cite{radford2021learning}. Note that we are not chasing high sample quality here, but using fast optimization to get results. More samples are in the appendix. DALL.E \cite{ramesh2021zero} sample are from their blog. Thank the authors. }\label{fig:t2i_sota}
\end{figure*}

\subsection{Exploiting Distribution Loss}

We point out in the Remark of Section \ref{sec:extra_loss} that, thanks to the scalability of our method, any potential loss function can be introduced to achieve some purpose.
In state-of-the-art methods (such as GANs and DDPMs), the distribution of the training set is implicitly exploited.
Given that it is reasonable to obtain the data distribution, we feed the data of each class to the neural network classifier, calculate the distribution of its output features, and use these statistics to construct a loss function to guide the generation of samples.
Specifically, let the mean of the feature of the $c$-th class data be $\mu_c$ and the variance be $\sigma_c^2$, then the distribution loss for generating samples of this class is formulated as:\begin{equation}\label{eqn:fid}
\begin{aligned}
&\mathcal{L}_\mathrm{distribution} (x^1, \cdots, x^N)\\
= & (\bar{x}_c - \mu_c)^T (\bar{x}_c - \mu_c) + (S_c - \sigma_c)^T (S_c - \sigma_c),
\end{aligned}
\end{equation}where:
\begin{equation}\label{eqn:stat}
\begin{aligned}
& \bar{x}_c = \frac{1}{N}\sum\limits_{i=0}^{N}
h(g(x^i)),\\
& S_c = \frac{1}{N-1} \sum\limits_{i=0}^{N}(h(g(x^i))-\bar{x}_c)^2.
\end{aligned}
\end{equation}Introducing this formula can significantly improve the diversity of samples generated by CaG.

\textbf{Remark.} Being able to properly optimize the input without introducing adversaries is one of the main properties of our method. This property makes our method highly scalable, opening up many possibilities for future applications.

\subsection{Sampling}\label{sec:sampling}

Sampling is the inference phase of a generative model. A good sampling strategy is critical to the sample generation quality.
According to the framework of our method introduced above, the role of random masking to CaG is roughly equivalent to Gaussian sampling to GANs and DDPMs, which is indispensable. The sampling process of CaG is detailed in Figure \ref{fig:method} (a), which looks very like the sampling process of DDPM. Specifically, although neural network classifiers and DDPMs are trained in quite different ways, they share some similarities in the sampling phase: they generate images by stepwise sampling, predict the sampling direction, and introduce noise to this direction.

\subsection{Text-to-Image Generation}\label{sec:t2i}

CaG can be easily generalized to text-to-image generation tasks.
As mentioned in Section \ref{sec:cls}, text-to-image foundation models (e.g., CLIP \cite{radford2021learning}) can be seen as a generalized classifier.
We can extract embeddings for text via the text encoder and combine the text embedding to form the weights of a linear layer, and then impose this linear layer on top of the image encoder. Then, we can get a neural network classifier. After obtaining this neural classifier, we can use Eqn. \eqref{eqn:mask} and either Eqn. \eqref{eqn:div} or Eqn. \eqref{eqn:fid} to produce realistic samples.

\section{Results}

\textbf{Datasets.} To evaluate our method, we conduct experiments on ImageNet at a scale of 256$\times$256 resolution.
I also experimented with text-to-image generation in the wild.

\textbf{Metrics.} We use the FID metric \cite{heusel2017gans} since it measures both the fidelity and diversity of the generated samples.
We also use the IS metric \cite{salimans2016improved} that measures how the generated instances are reasonable and how they capture the distribution of the overall dataset.
While these metrics are, in many cases, good indicators of the sample quality
, they are still problematic. \textbf{So we also show a lot of pictures.}

\begin{table}[t]
\footnotesize
\centering
\caption{Quantitative comparison between our method and state-of-the-art methods on the ImageNet 256$\times$256 task. $\dagger$ indicates a differentiable FID score.
}
\begin{tabular}{c|c|c|c} 
 \hline
\toprule
Resolutions& Methods & FID $\downarrow$  & IS $\uparrow$\\

 \midrule
\multirow{12}{*}{$256\times 256$} & BigGAN-deep \cite{brock2018large} & 6.95 & \\

& IDDPM \cite{nichol2021improved} & 12.26  &  \\
& SR3 \cite{saharia2022image} &  11.30 &    \\

& DCTransformer \cite{nash2021generating} &  36.51 &\\
& VQ-VAE-2 \cite{razavi2019generating} & 31.11 &\\
& ADM \cite{dhariwal2021diffusion} w/o condition, w/ guidance & 12.00 & 95.41 \\

& DeepDream \cite{Mordvintsev2016DeepDream} & 134.69 & 22.60 \\
& Ours & 6.88 $\dagger$  & 326.33 \\ 

\bottomrule
\end{tabular}
\label{tab:sota}
\end{table}

\subsection{State-of-the-Art Image Synthesis}

The table compares CaG with state-of-the-art techniques in Table \ref{tab:sota}. Our method achieves promising numbers, being comparable to DDPMs and GANs. Note that these comparisons are a bit unfair for our methods.
This is because some of the competitors employ super-resolution modules that we did not engage in, which is unfair to our method and leaves our approach at an unfair disadvantage position.
Regardless, our method achieves good numbers, and it does not require complex training as prior generative models do.

\begin{figure*}[t]
\centering
\includegraphics[width=1.0\linewidth]{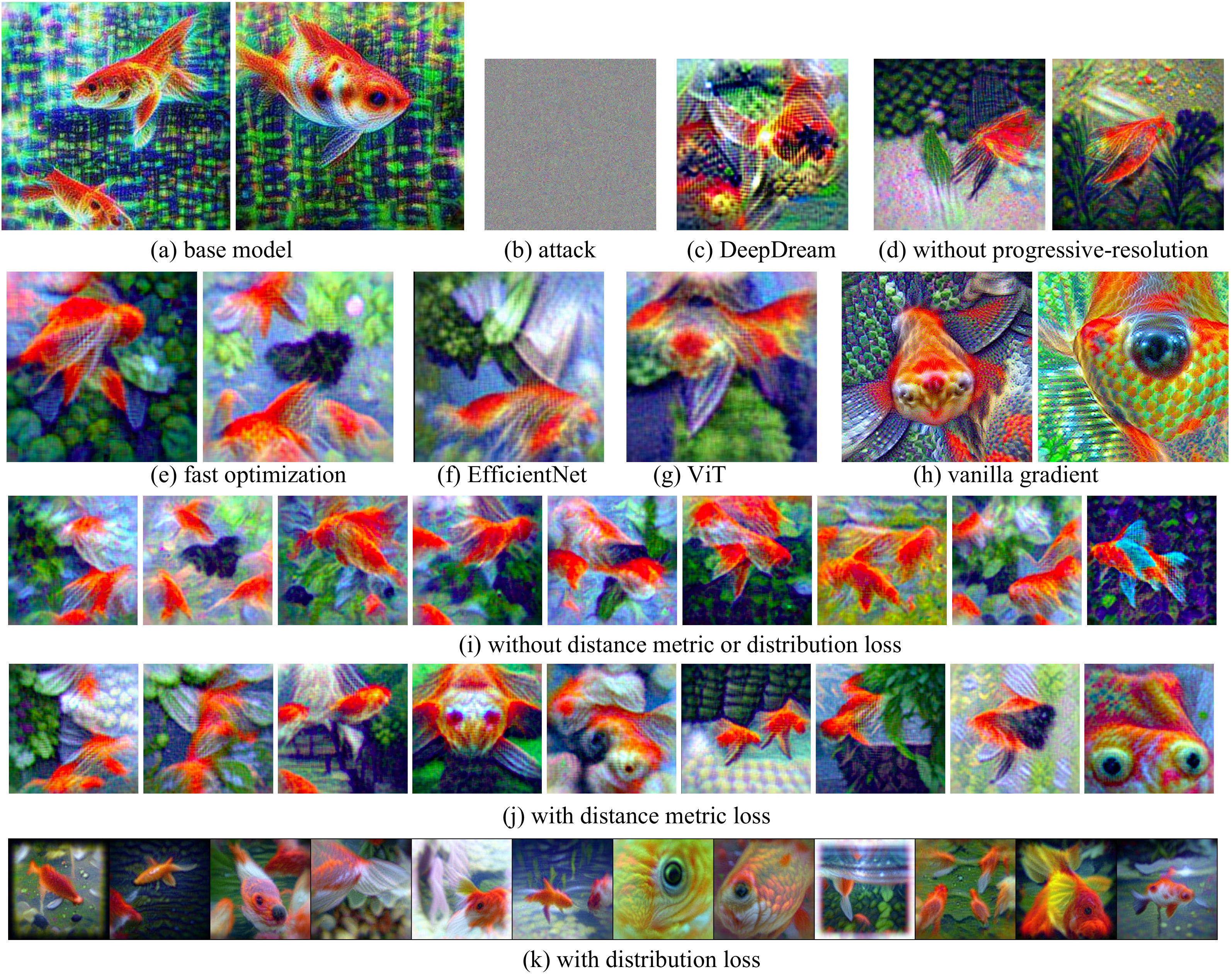}
\vspace{-11pt}
\caption{Ablation studies on our method. Note that in order to obtain the results quickly, we use fast optimization (i.e., fewer sampling steps with gradient blurring) except for Figures \ref{fig:ablation} (a) and (g). Therefore, the resolution of each image is somewhat low.}\label{fig:ablation}
\vspace{-11pt}
\end{figure*}

Figure \ref{fig:sota} compares randomly selected generated samples of CaG with that of best-performing methods, including BigGaN-deep \cite{brock2018large}, ADM \cite{dhariwal2021diffusion}, and DeepDream \cite{Mordvintsev2016DeepDream}.
The three represent state-of-the-art GAN, DDPM, and feature visualization algorithms, respectively.
In terms of visual effects, all methods have good visual perceptual quality except DeepDream, which has very poor visual quality.
However, our method has stronger semantic perception, resulting in more significant semantic-driven diversity than GANs and DDPMs.
\textbf{First}, CaG pays more attention to object diversity than background diversity, proving that CaG is more semantically aware of diversity. 
For example, in images generated by CaG, birds occupy a large area of the picture.
This is consistent with human cognition, for background diversity is easy to achieve by pasting objects to different backgrounds, while object diversity is challenging to obtain.
\textbf{Second}, CaG decouples and removes irrelevant object categories, generating only the categories we want.
For example, CaG's samples frequently included only individual Tinca fishes that were not held by people, significantly different from the samples of GANs, DDPMs, and even the training set.
\textbf{Third}, the CaG appears to be aware of geometric information. The images generated by CaG have different orientations, and these orientations include not only the affine transformation of the plane but also different views in the 3D perspective.
Finally, CaG is more reasonable, high-fidelity, and diverse than traditional feature visualization methods. Moreover,the CaG algorithm is more robust and reproducible than DeepDream and does not require excessive parameter tuning.

We have provided samples generated by our method in Figure \ref{fig:intro} and Figure \ref{fig:sota} of the paper. However, the optimization time for each sample needs to be longer, and the number of samples shown needs to be more. Therefore, we present a large number of samples in detail in this supplementary material. Please feel free to review Figures \ref{fig:supp1}, \ref{fig:supp2}, \ref{fig:supp3}, \ref{fig:supp4}, \ref{fig:supp5}, \ref{fig:supp6}, \ref{fig:supp7}, \ref{fig:supp8}, \ref{fig:supp9}, \ref{fig:supp10}, and \ref{fig:supp11} below in the Appendix.

\subsection{Text-to-Image Generation}

As described in Section \ref{sec:t2i}, our method can be used for text-to-image generation straightforwardly. We downloaded the CLIP model \cite{radford2021learning} and treated CLIP as a generalized classifier, as described earlier. We then use Eqn. \eqref{eqn:mask} for optimization. The results are presented in Figure \ref{fig:t2i}.

Since this is just an extended experiment, we do not pursue the quality of the generated samples, so a quick experiment is performed. Figure \ref{fig:t2i} shows that our method has great potential for text-to-image generation.
Our method can not only understand abstract semantics for image generation (such as ``a man loves a woman") but also generate scenarios that do not exist in reality (such as ``a man plays basketball on the moon"), which is very intriguing.

We also compare our method with the state-of-the-art text-to-image generation method DALL.E \cite{ramesh2021zero}. As shown in Figure \ref{fig:t2i_sota}, our method generates images with higher fidelity than DALL.E. Be aware that our method is just a generic classifier that is not specifically trained for generative purposes.

In the appendix, we show more examples of text-to-image generation. Please feel free to review Figures \ref{fig:i2t_supp1}, \ref{fig:i2t_supp2}, \ref{fig:i2t_supp3}, and \ref{fig:i2t_supp4} below in the Appendix.

\subsection{Ablation Study}

In this section, we analyze the components of our algorithm to provide deeper insights.

\textbf{Mask-Based Stochastic Reconstruction.} It is widely known that when a target is given, optimizing the input is like a targeted adversarial attack -- although the output prediction of the system can be the same as the given target, human-meaningful samples cannot be generated.
This is because the partial derivative of the loss function with respect to the input is semantic-agnostic.
Here, we examine whether an additional reconstruction module can regularize the partial derivatives to make them semantic-aware to produce human-meaningful images \cite{wang2022semantic,he2022masked,sabour2017dynamic}.
Figure \ref{fig:ablation} (a) is the base model, and Figure \ref{fig:ablation} (b) is a model similar to Figure \ref{fig:ablation} (a) but without a mask-based stochastic reconstruction module.
The comparison between Figure \ref{fig:ablation} (a) and (b) showcase this reconstruction module plays a critical role in producing human-meaningful samples.

DeepDream \cite{Mordvintsev2016DeepDream} avoids the semantic-agnostic gradient problem by smoothing the gradients with Gaussian blurring. We also compare the mask-based stochastic reconstruction technique with gradient blurring in Figure \ref{fig:ablation} (a) and (c). The results show that our method yields remarkably better results than gradient blurring.
These comparisons prove the indispensability of the stochastic reconstruction module, justifying our conjecture.

\textbf{Progressive-Resolution.} Stepwise sample generation from 64$\times$64 to 256$\times$256 resolution is adopted as our strategy. For comparison, here we remove this progressive resolution strategy, i.e., we directly generate images of the desired resolution. The experimental results are shown in Figure \ref{fig:ablation} (d), from which we can see that after removing the progressive resolution strategy, some strange patterns appear in the sample, and the sample fidelity is reduced, which proves the necessity of the progressive resolution strategy.

\textbf{Distance metric loss.} Besides classification loss, we also introduce distance metric loss to encourage the diversity of samples. Figure \ref{fig:ablation} (j) contains distance metric loss while \ref{fig:ablation} (i) does not contain it.
The comparison between Figure \ref{fig:ablation} (i) and (j) shows that distance metric loss can indeed increase the variety of samples.
It produces samples with greater sematic-aware diversity than cross-entropy loss and does not mechanically mimic the training set as cross-entropy loss does. This verifies the effectiveness of the proposed distance metric loss.

\textbf{Distribution Loss.}
Next, we ablate the distribution loss for the study. By comparing Figures \ref{fig:ablation} (i) and (k), we can confirm that distribution loss can improve the diversity and fidelity of samples. Further, by comparing Figures \ref{fig:ablation} (j) and (k), we can know that the effects of distribution loss and distance metric loss are slightly different.

\textbf{Architectural Impact.} Unlike prior generators that required difficult training, we do not need to train any models. Our neural network classifier is downloaded from the internet \footnote{\url{https://pytorch.org/vision/stable/models.html}; \url{https://github.com/rwightman/pytorch-image-models}.}, and the mask-based stochastic reconstruction module is also downloaded from the internet \footnote{\url{https://github.com/facebookresearch/mae}; \url{https://github.com/open-mmlab/mmselfsup}.}.
Given that so many models are available on the Internet, we are free to choose any architecture for image generation.
The comparison between Figure \ref{fig:ablation} (f) and (g) shows that different architectures generate different samples.
Here, without loss of generality, we choose two typical neural network classifier architectures: Inception \cite{szegedy2016rethinking} as the representative of convolutional neural networks \cite{lecun1998gradient}, and ViT-S \cite{dosovitskiy2020image} as the representative of vision transformers \cite{dosovitskiy2020image}.
For simplicity, we choose ViT-B for the mask-based random reconstruction module to save computational costs.

We also try to integrate neural network classifiers of different architectures. Specifically, we optimize an input such that it can reduce the loss of Inception and ViT-S simultaneously.
The comparison of Figure \ref{fig:ablation} (e) vs. (f) and the comparison of Figure \ref{fig:ablation} (e) vs. (g) show that ensembling can further improve the quality of the generated images. Therefore, in this paper, we use this ensembling setting by default.

\textbf{Sampling Steps.} Here, we compare the samples at different sampling steps to determine the relationship between image generation quality and sampling steps. Figure \ref{fig:ablation} (a) shows that the best generation results are achieved when the sampling step number is about 3000. 
Most of the time, in order to quickly verify our algorithm, we only sample 2000 times (see ``fast optimization" in  Figure \ref{fig:ablation} (e)). Increasing the sampling steps within a certain range will make the effect better. Increasing sampling steps further might result in over-optimization.

\textbf{Gradient Blurring.} Prior arts \cite{Mordvintsev2016DeepDream} showed that Gaussian blurring of gradients is beneficial to increase their semantic perception. Therefore, we also explored the comparison of adding gradient blurring and not adding gradient blurring. 
Specifically, gradient blurring refers to filtering the gradients of the three channels separately with Gaussian filters. Comparing Figures \ref{fig:ablation} (e) and (h), we can find that using gradient blur makes the optimization converge faster, but at the expense of image fidelity.
Therefore, in the case of fast optimization, gradient blurring is used by default in this paper. But it must be made clear that in our method, gradient blurring is not required.

\section{Conclusion}

We showcase that traditional classification neural networks have strong image generation capabilities, which were previously overlooked.
Proving that classifiers have learned the data distribution and are ready for image generation has far-reaching implications, for classifiers are much easier to train than generative models like DDPMs and GANs. We don't even need to train classification models because tons of public ones are available for download. Also, this holds great potential for the interpretability and robustness of classifiers. For example, people have wondered what explainable artificial intelligence could look like for years. In particular, efforts are being made to try to build new interpretable neural networks. 
Given that traditional classification neural networks can be used to generate diverse and high-fidelity samples, they can also be used to produce explanations for their decision-making or their predictions.
We conjecture that new neural networks may not be needed because existing neural networks already have sufficient interpretability.

\textbf{Broader Impacts and Limitations.} Proving that the neural network classifier can complete the generation task can make classification neural networks more transparent but may also bring some unpredictable and uncontrollable privacy or data leakage problems.

{\small
\bibliographystyle{ieee_fullname}
\bibliography{egbib}
}

\clearpage

\begin{appendix}

\twocolumn[{%
\renewcommand\twocolumn[1][]{#1}%
\maketitle
\begin{center}
    \centering
    \textbf{\Huge{Supplementary Materials of ``Traditional Classification Neural Networks are Good Generators''\\}}
    \vspace{22pt}
    \large{Guangrun Wang and Philip H.S. Torr\\}
    \vspace{11pt}
    \large{University of Oxford\\}
    \vspace{11pt}
    \large{
    \{guangrun.wang, philip.torr\}@eng.ox.ac.uk
    }
    \vspace{44pt}
\end{center}%
}]

We provide a number of samples generated by CaG in this appendix, and we very much look forward to the readers viewing the images in the appendix below.

\section{Samples of CaG on ImageNet 256$\times$256}

We have provided samples generated by our method in Figure \ref{fig:intro} and Figure \ref{fig:sota} of the paper. However, the optimization time for each sample needs to be longer, and the number of samples shown needs to be more. Therefore, we present a large number of samples in detail in this supplementary material.

\begin{figure*}
\centering
\includegraphics[width=1.0\linewidth]{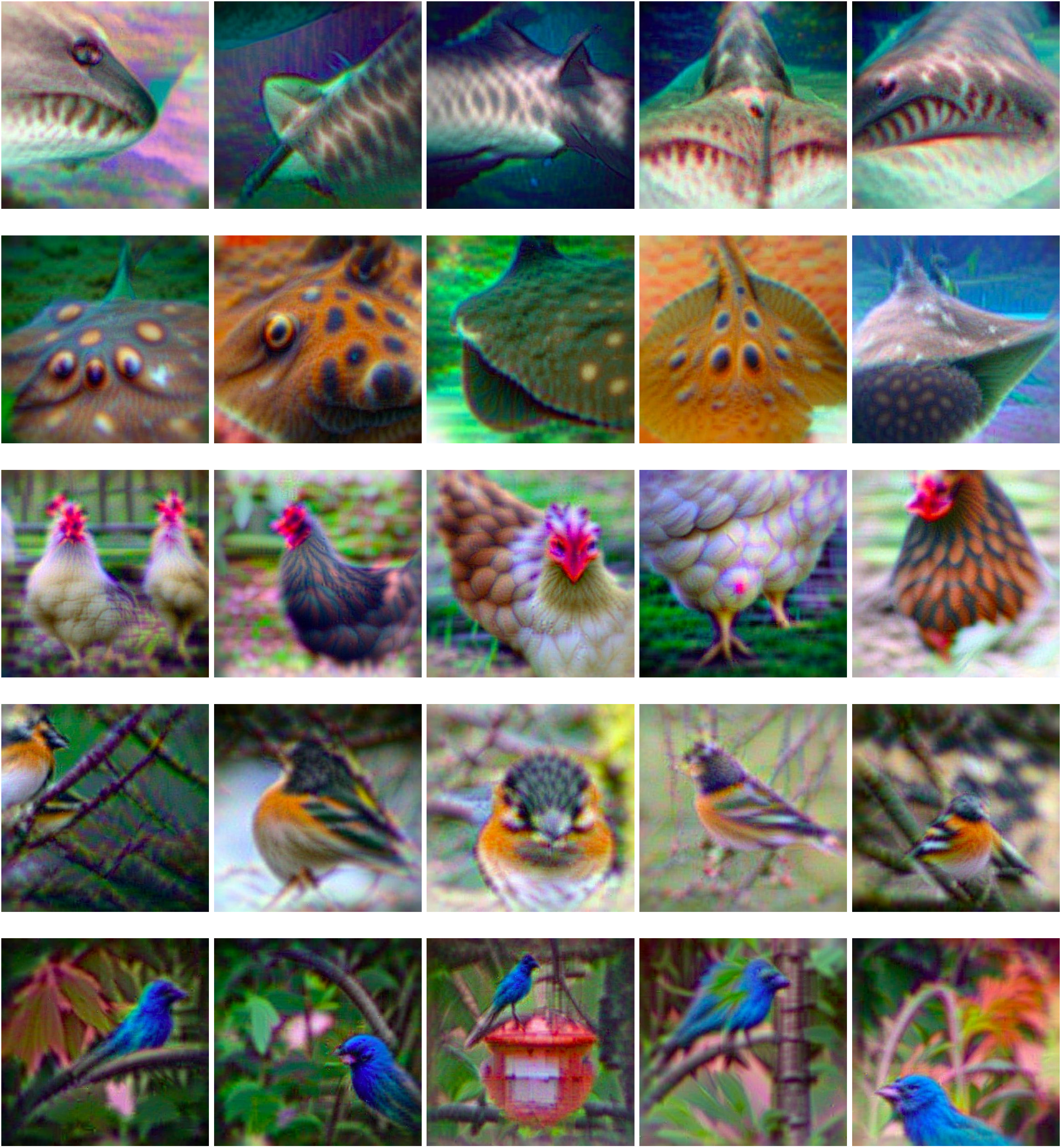}
\caption{Samples of CaG on ImageNet 256$\times$256. The classes are Class 3 (i.e.,  tiger shark), Class 5 (i.e., electric ray), Class 8 (i.e., hen), Class 10 (i.e., brambling), and Class 14 (i.e., indigo bunting).}\label{fig:supp1}
\end{figure*}

\begin{figure*}
\centering
\includegraphics[width=1.0\linewidth]{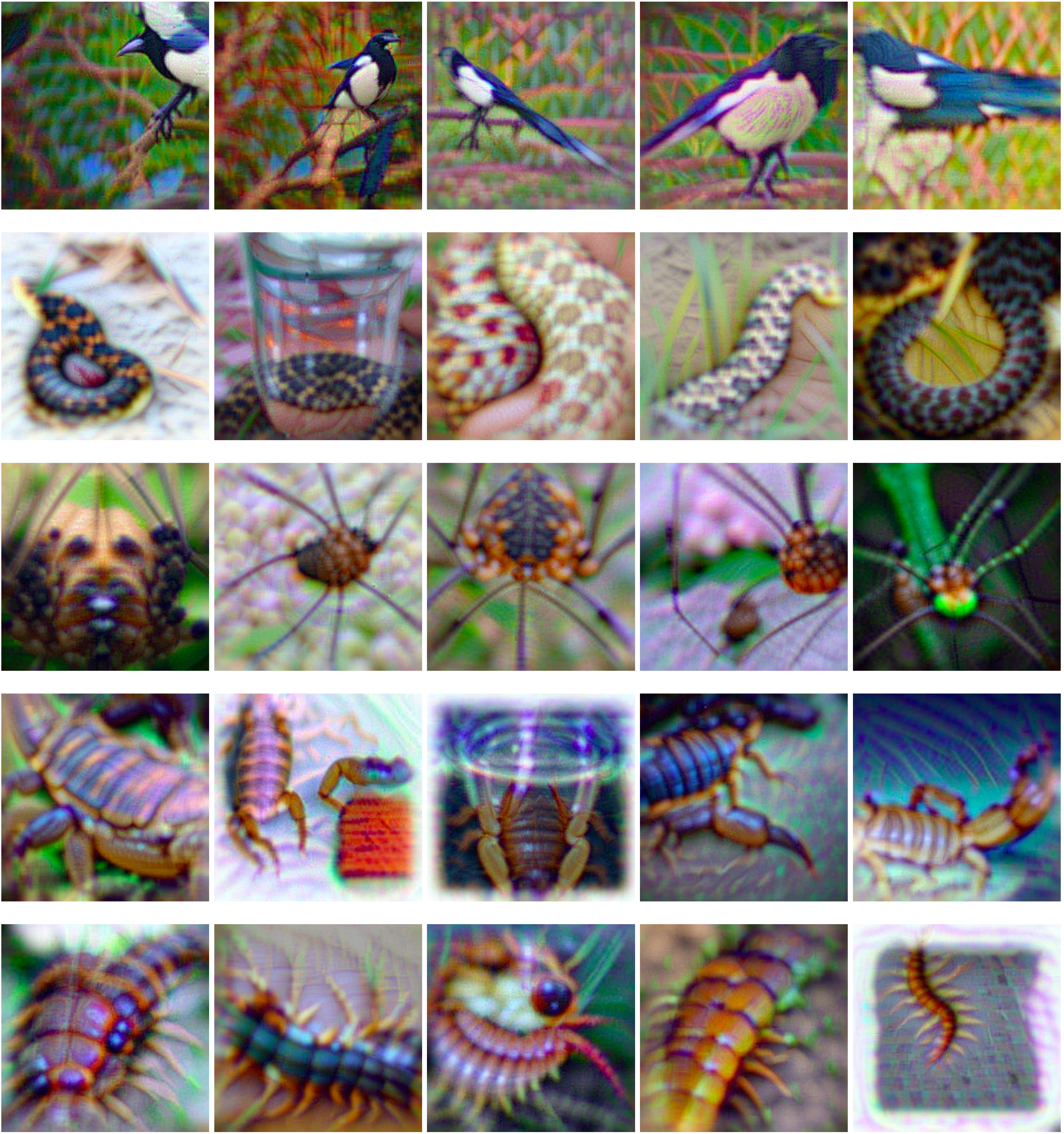}
\caption{Samples of CaG on ImageNet 256$\times$256. The classes are Class 18 (i.e., magpie), Class 54 (i.e., hognose snake), Class 70 (i.e., harvestman), Class 71 (i.e., scorpion), and Class 79 (i.e., centipede).}\label{fig:supp2}
\end{figure*}

\begin{figure*}
\centering
\includegraphics[width=1.0\linewidth]{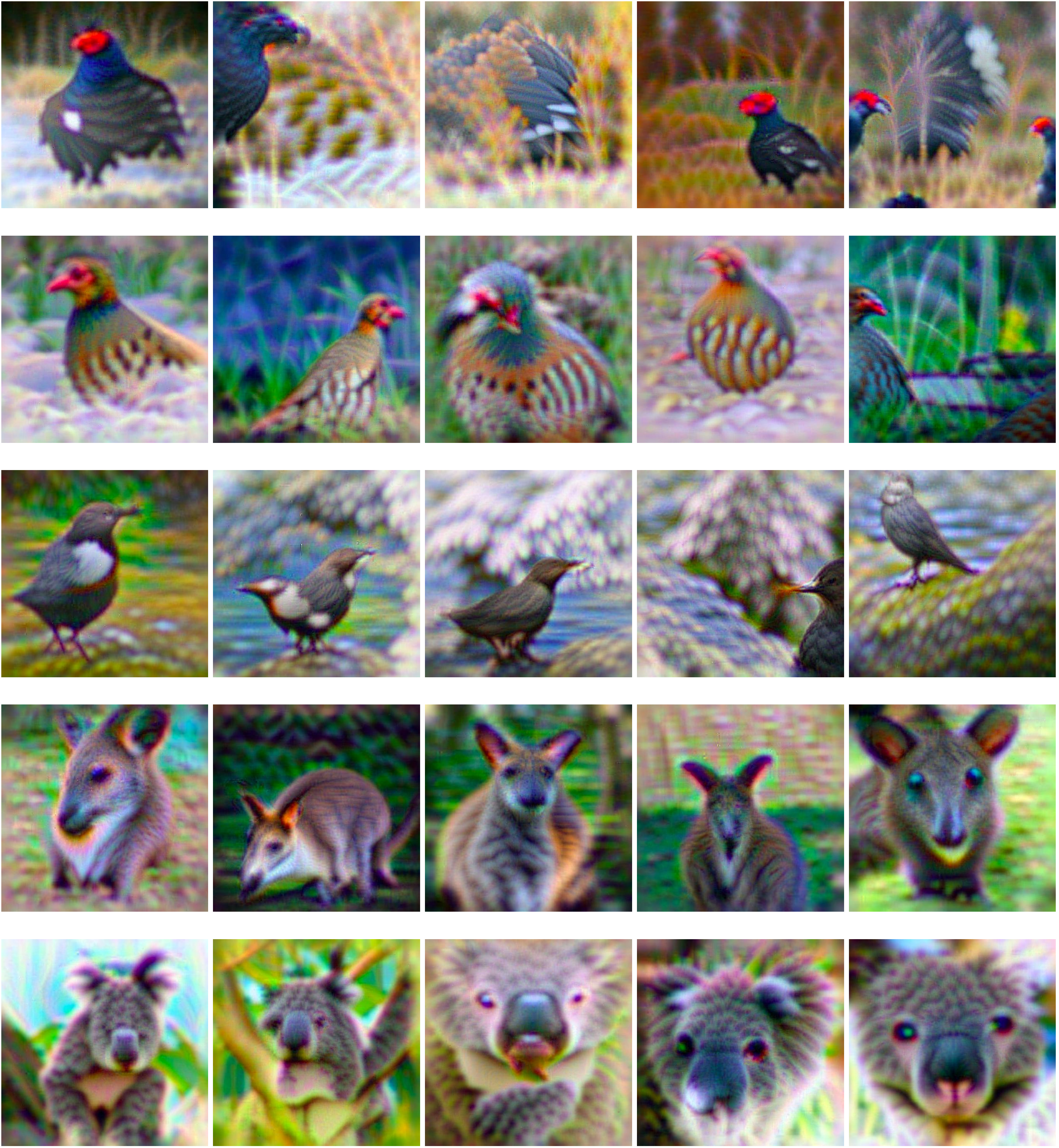}
\caption{Samples of CaG on ImageNet 256$\times$256. The classes are Class 80 (i.e., black grouse), Class 86 (i.e., partridge), Class 20 (i.e., water ouzel), Class 104 (i.e., wallaby), and Class 105 (i.e., koala).}\label{fig:supp3}
\end{figure*}

\begin{figure*}
\centering
\includegraphics[width=1.0\linewidth]{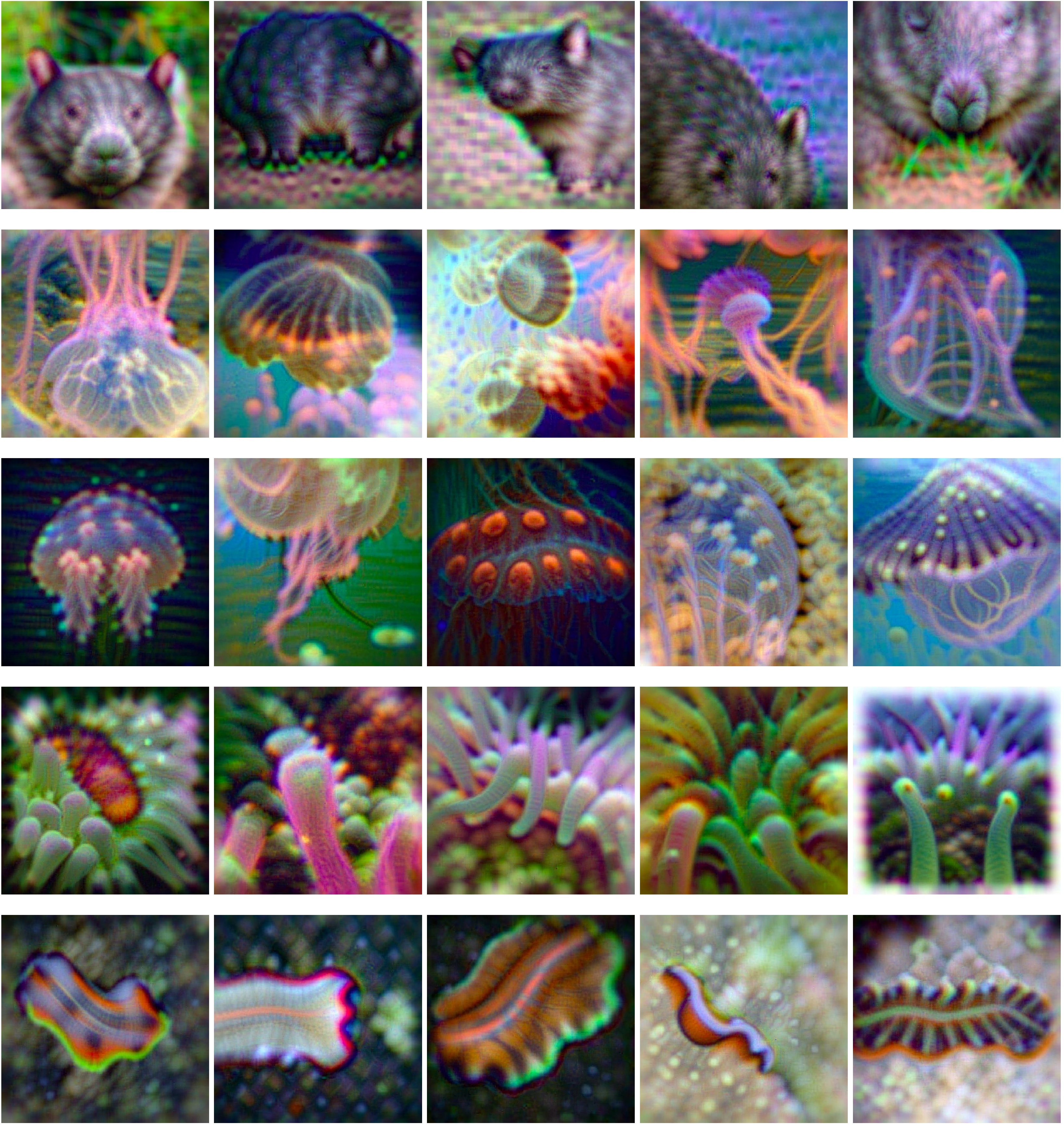}
\caption{Samples of CaG on ImageNet 256$\times$256. The classes are Class 106 (i.e., wombat), Class 107 (i.e., jellyfish), Class 107 (i.e., jellyfish), Class 108 (i.e., sea anemone), and Class 110 (i.e., flatworm).}\label{fig:supp4}
\end{figure*}

\begin{figure*}
\centering
\includegraphics[width=1.0\linewidth]{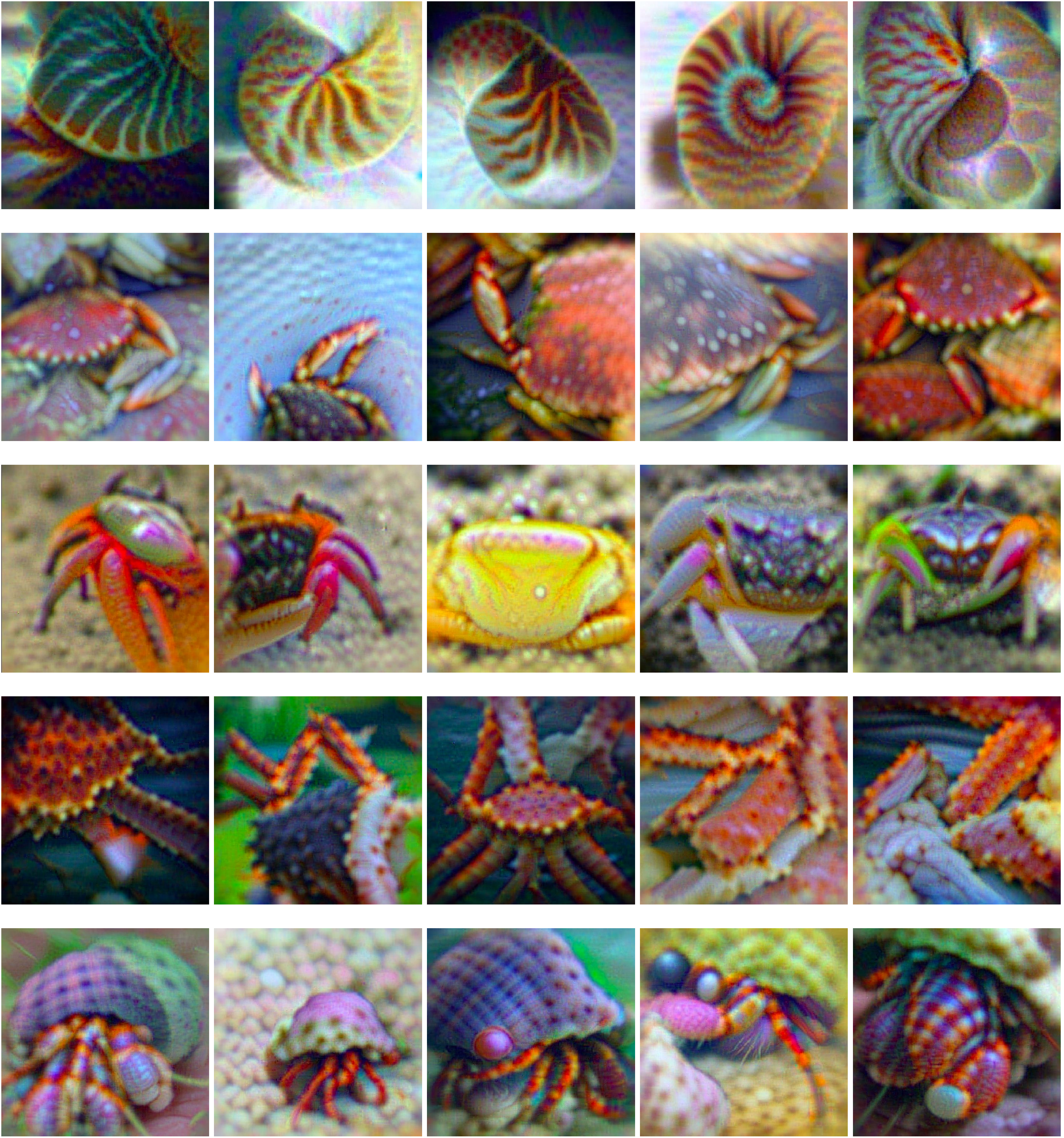}
\caption{Samples of CaG on ImageNet 256$\times$256. The classes are Class 117 (i.e., chambered nautilus), Class 118 (i.e., Dungeness crab), Class 120 (i.e., fiddler crab), Class 121 (i.e., king crab), and Class 125 (i.e., hermit crab).}\label{fig:supp6}
\end{figure*}

Please feel free to review Figures \ref{fig:supp1}, \ref{fig:supp2}, \ref{fig:supp3}, \ref{fig:supp4}, \ref{fig:supp5}, \ref{fig:supp6}, \ref{fig:supp7}, \ref{fig:supp8}, \ref{fig:supp9}, \ref{fig:supp10}, and \ref{fig:supp11} below in the Appendix.

\section{Samples of Text-to-Image Generation}

In Figure \ref{fig:t2i} of the article, we show that our method can not only use neural network classifiers for sample generation but, more excitingly, it can also be used for text-to-image generation. We can treat a text-image pre-training model as a generalized classifier and extend CaG to text-to-image generation. 
In this appendix, we show more examples of text-to-image generation.
Please note that we are not chasing high sample quality here, but using fast optimization to get results. We also did not use distance metric loss to increase the diversity of samples.

Please feel free to review Figures \ref{fig:i2t_supp1}, \ref{fig:i2t_supp2}, \ref{fig:i2t_supp3}, and \ref{fig:i2t_supp4} below in the Appendix.

\end{appendix}

\begin{figure*}
\centering
\includegraphics[width=1.0\linewidth]{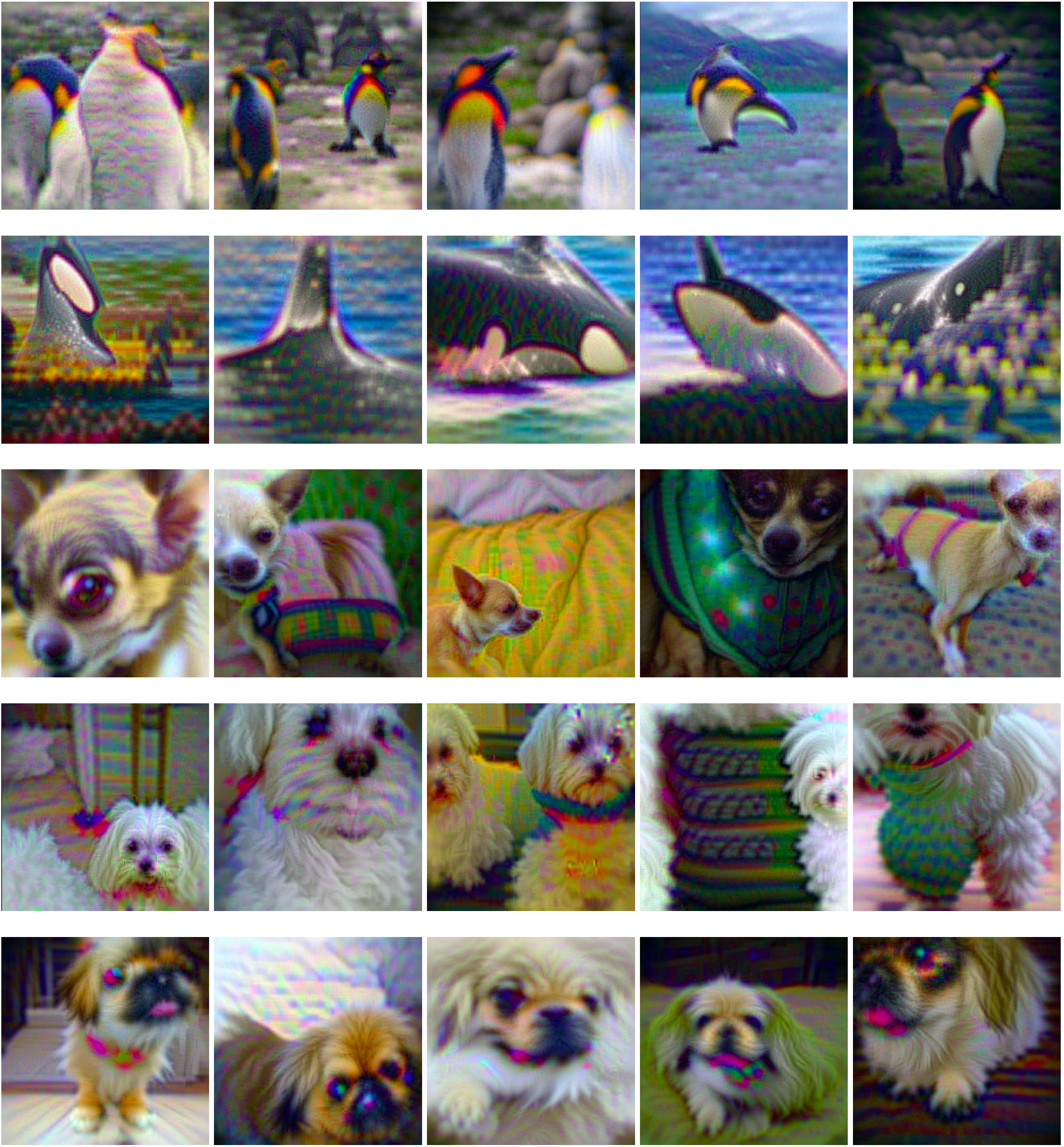}
\caption{Samples of CaG on ImageNet 256$\times$256. The classes are Class 145 (i.e., king penguin), Class 148 (i.e., killer whale), Class 151 (i.e., Chihuahua), Class 153 (i.e., Maltese dog), and Class 154 (i.e., Pekinese).}\label{fig:supp8}
\end{figure*}

\begin{figure*}
\centering
\includegraphics[width=1.0\linewidth]{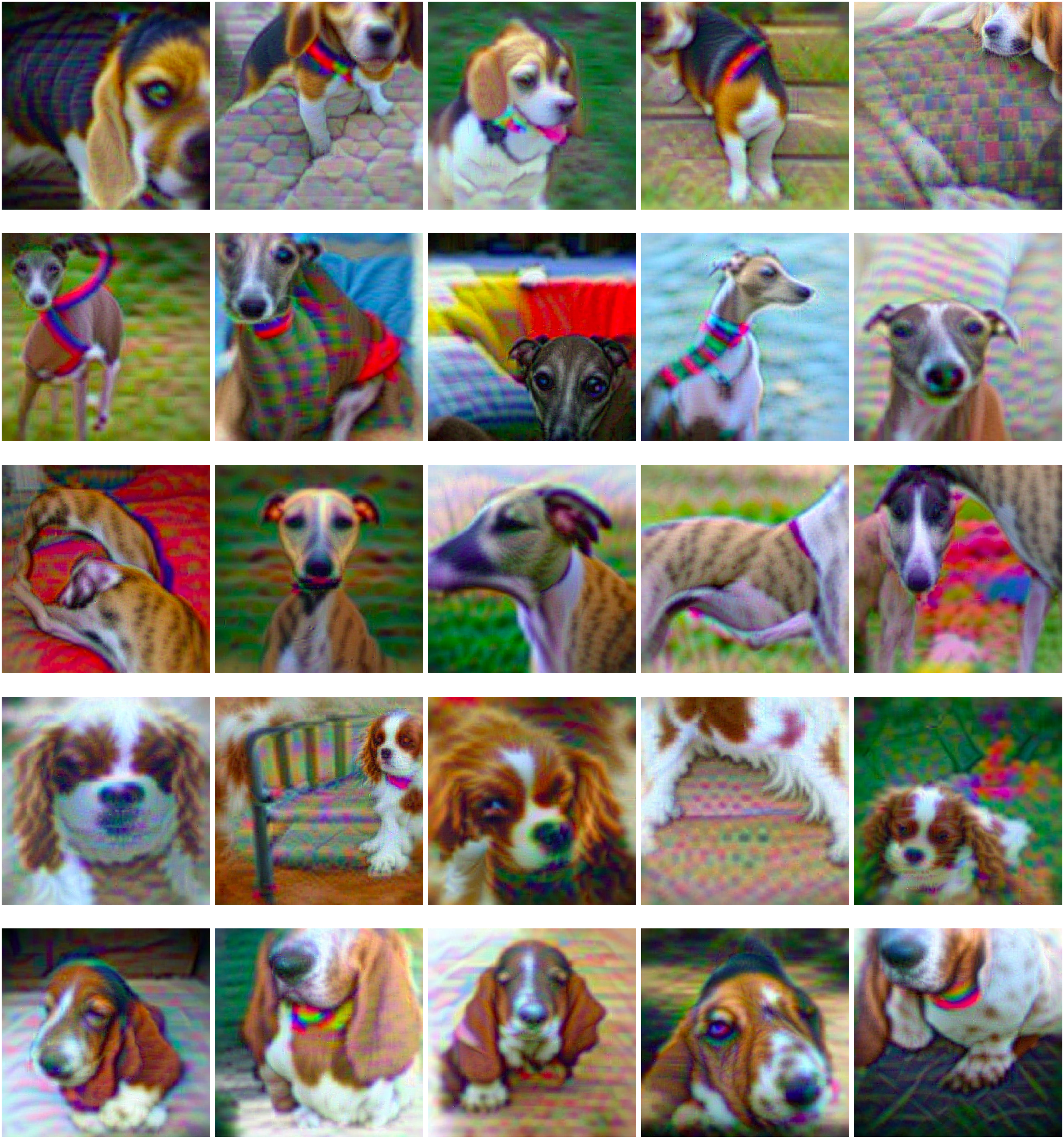}
\caption{Samples of CaG on ImageNet 256$\times$256. The classes are Class 162 (i.e., beagle), Class 171 (i.e., Italian greyhound), Class 172 (i.e., whippeta), Class 156 (i.e., Blenheim spaniel), and Class 161 (i.e., basset).}\label{fig:supp9}
\end{figure*}

\begin{figure*}
\centering
\includegraphics[width=1.0\linewidth]{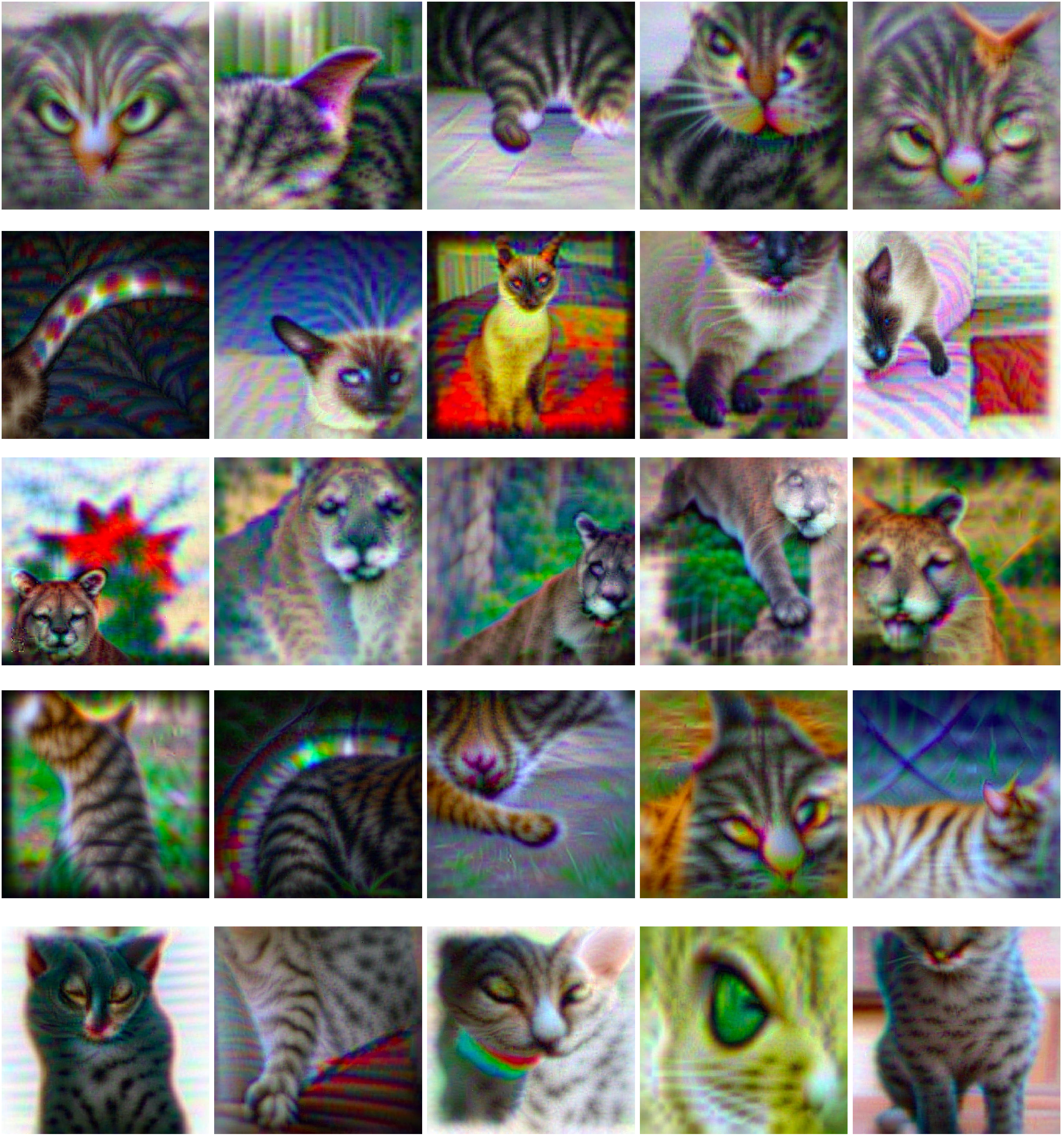}
\caption{Samples of CaG on ImageNet 256$\times$256. The classes are Class 281 (i.e., tabby cat), Class 284 (i.e., Siamese cat), Class 286 (i.e., cougar), Class 282 (i.e., tiger cat), and Class 285 (i.e., Egyptian cat).}\label{fig:supp10}
\end{figure*}

\begin{figure*}
\centering
\includegraphics[width=1.0\linewidth]{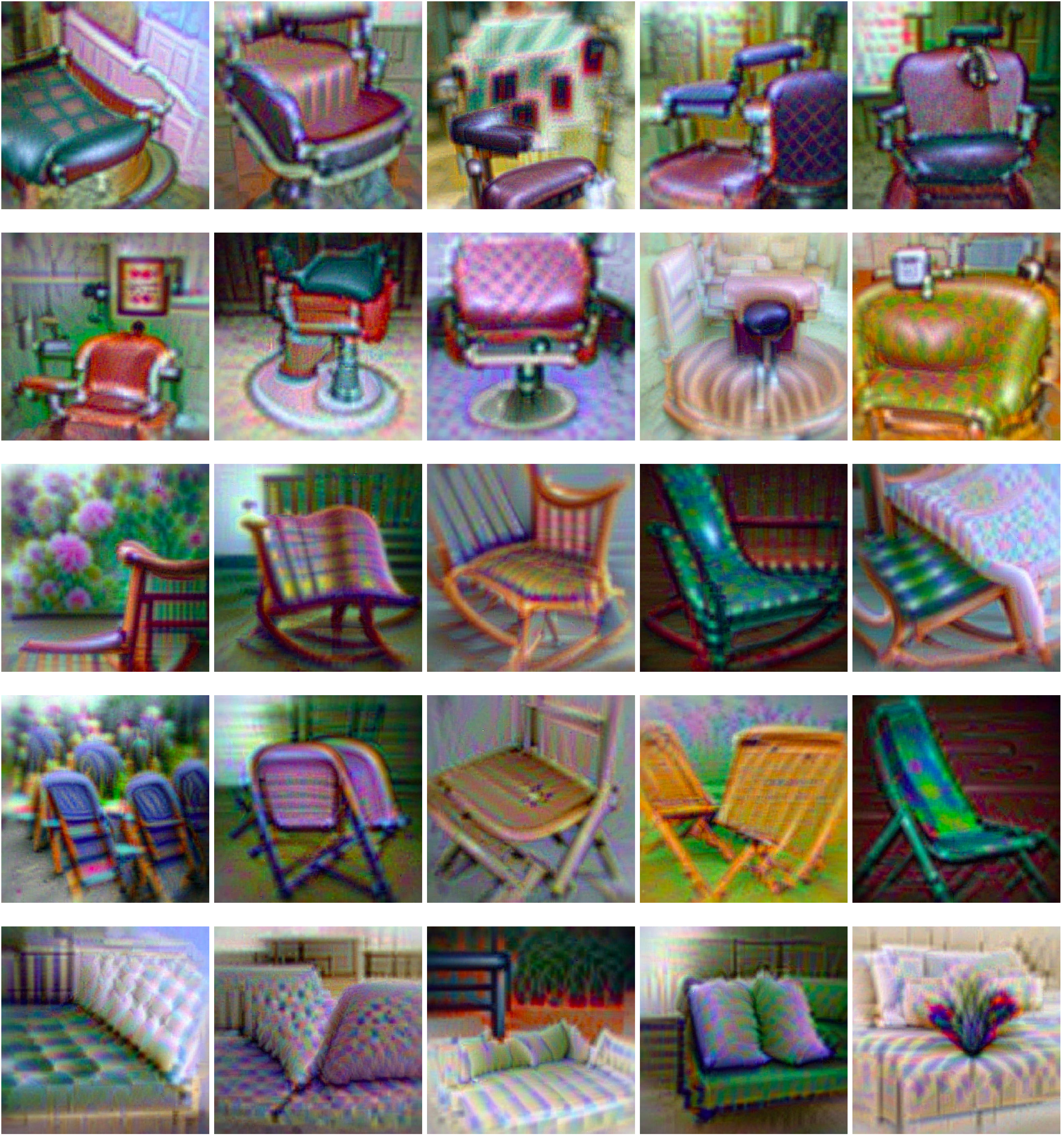}
\caption{Samples of CaG on ImageNet 256$\times$256. The classes are Class 423 (i.e., barber chair), Class 423 (i.e., barber chair), Class 765 (i.e., rocking chair), Class 559 (i.e., folding chair), and Class 831 (i.e., studio couch).}\label{fig:supp11}
\end{figure*}

\begin{figure*}
\centering
\includegraphics[width=1.0\linewidth]{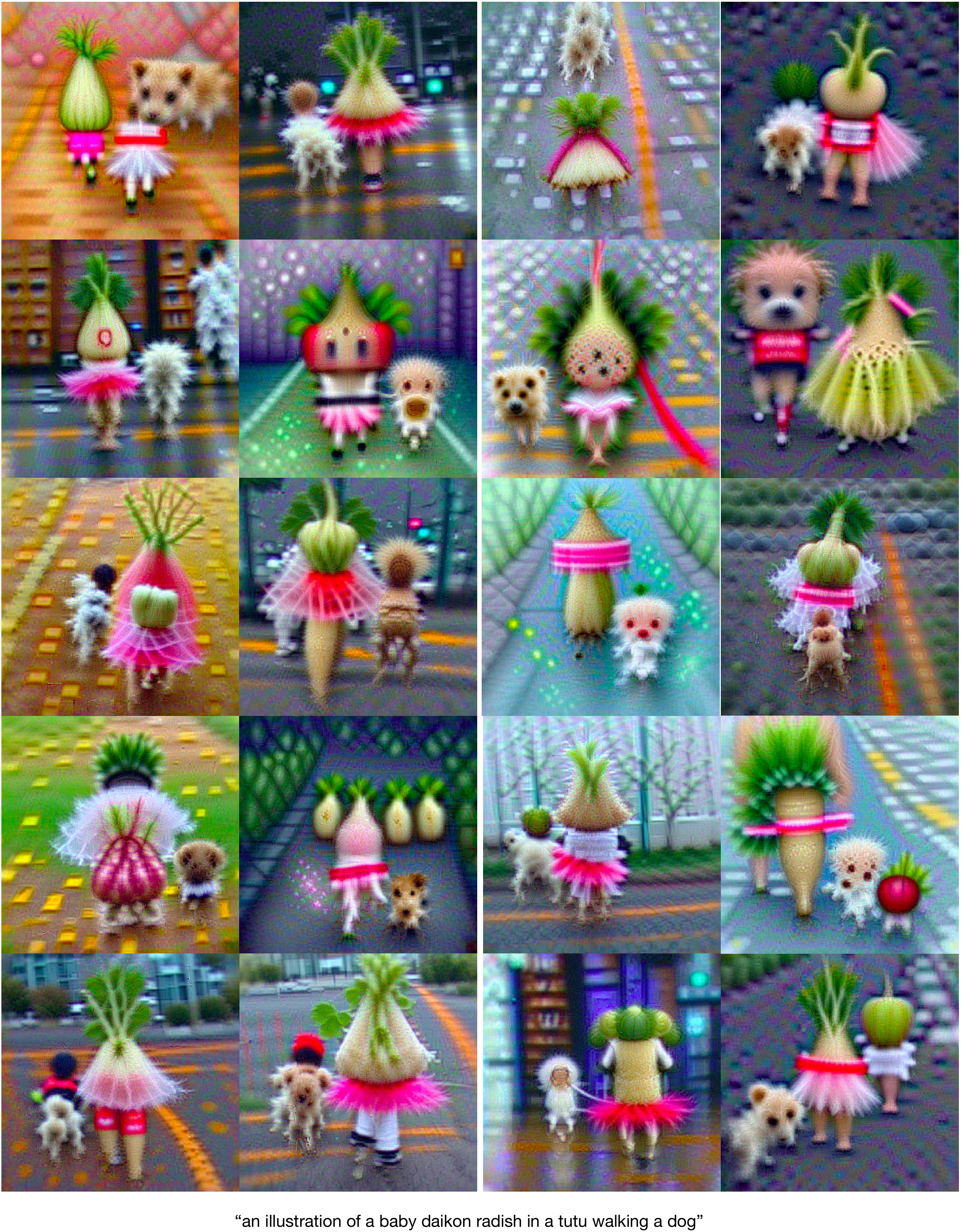}
\caption{Text-to-image generation using a pretrained text-image model CLIP. Note that we are not chasing high sample quality here, but using fast optimization to get results. We also did not use distance metric loss to increase the diversity of samples. }\label{fig:i2t_supp1}
\end{figure*}

\begin{figure*}
\centering
\includegraphics[width=1.0\linewidth]{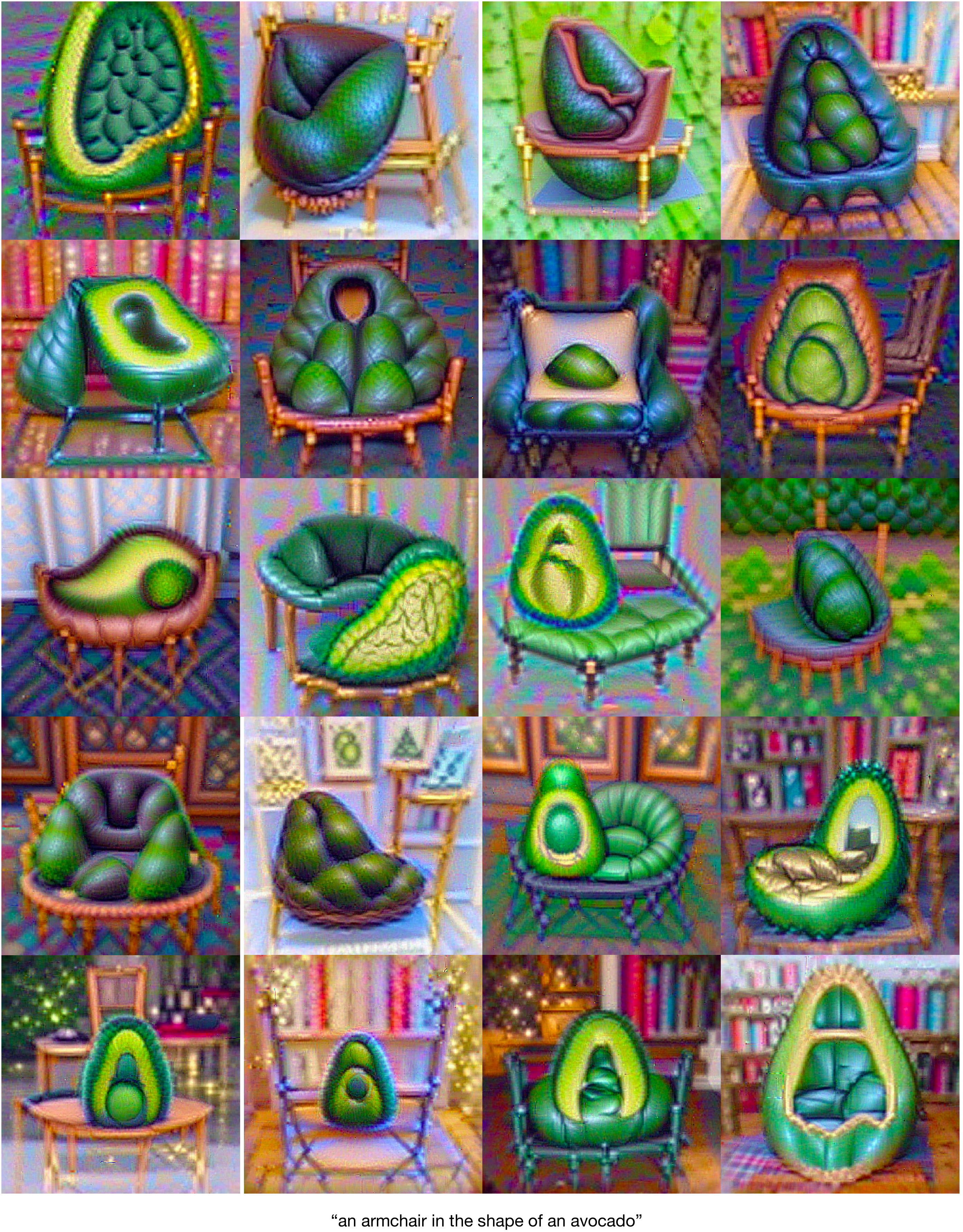}
\caption{Text-to-image generation using a pretrained text-image model CLIP. Note that we are not chasing high sample quality here, but using fast optimization to get results. We also did not use distance metric loss to increase the diversity of samples. }\label{fig:i2t_supp2}
\end{figure*}

\begin{figure*}
\centering
\includegraphics[width=1.0\linewidth]{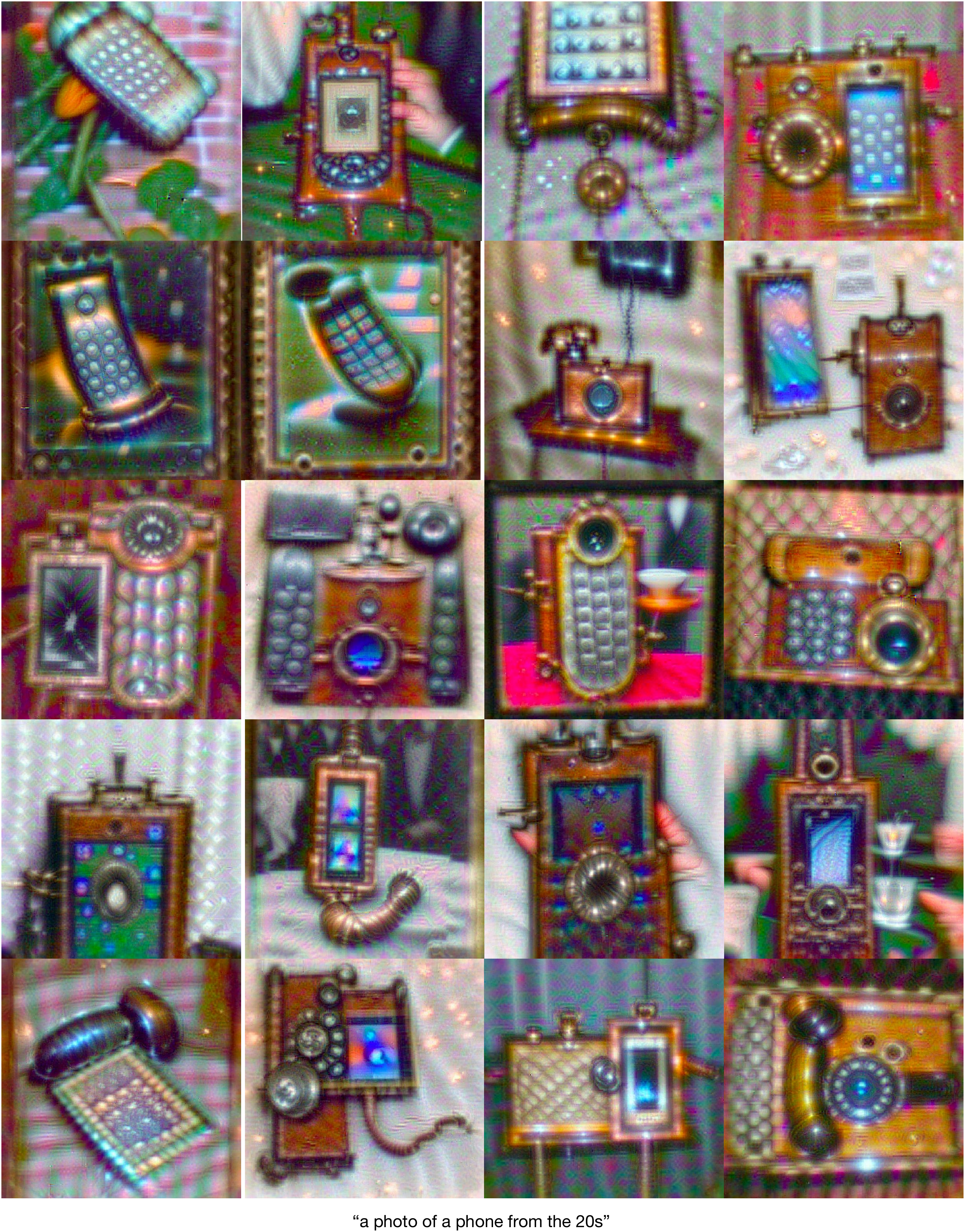}
\caption{Text-to-image generation using a pretrained text-image model CLIP. Note that we are not chasing high sample quality here, but using fast optimization to get results. We also did not use distance metric loss to increase the diversity of samples. }\label{fig:i2t_supp3}
\end{figure*}

\begin{figure*}
\centering
\includegraphics[width=1.0\linewidth]{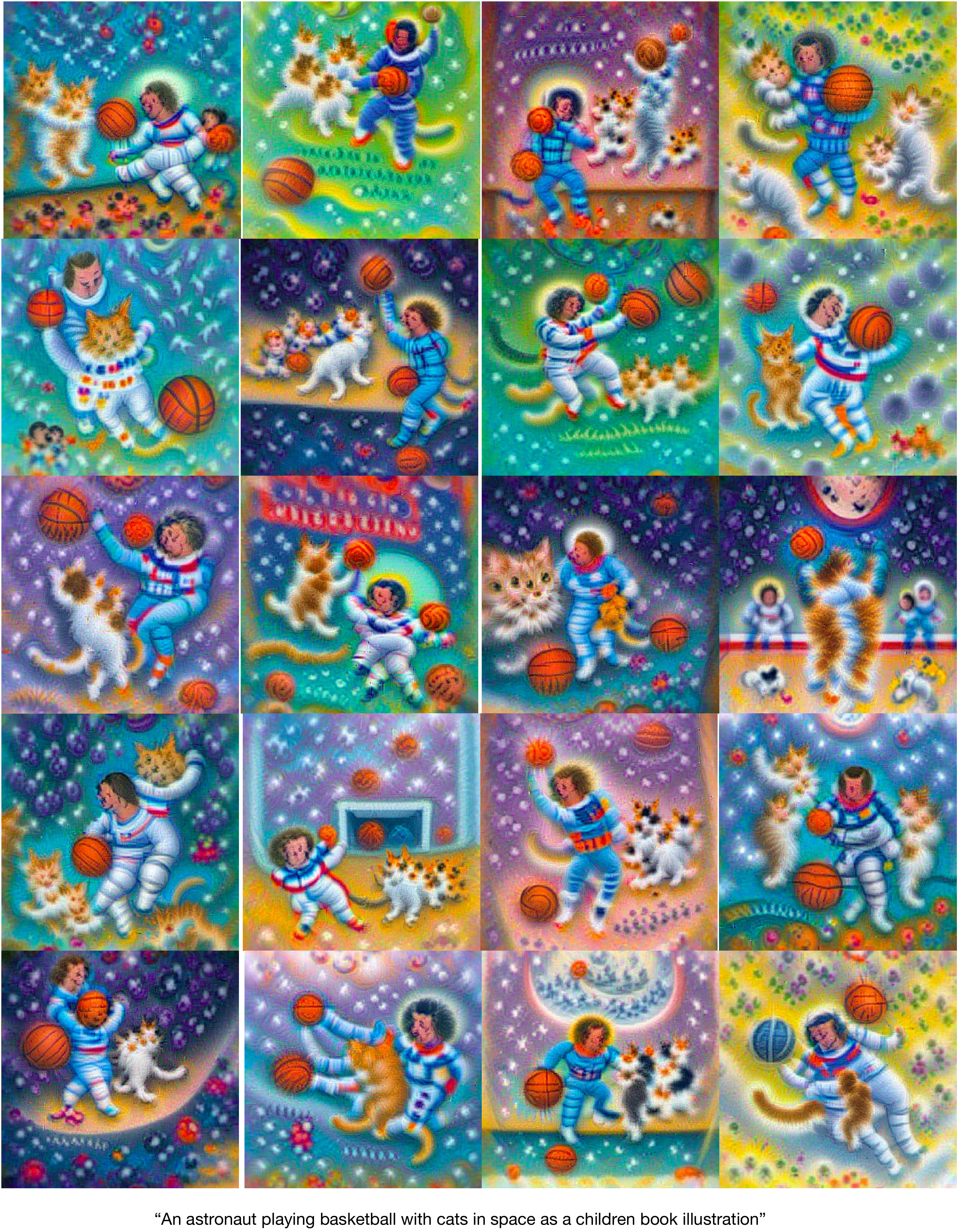}
\caption{Text-to-image generation using a pretrained text-image model CLIP. Note that we are not chasing high sample quality here, but using fast optimization to get results. We also did not use distance metric loss to increase the diversity of samples. }\label{fig:i2t_supp4}
\end{figure*}


\end{document}